\documentclass[11pt]{article}

\usepackage[final]{acl}

\usepackage{times}
\usepackage{latexsym}
\usepackage{adjustbox}
\usepackage{multirow}

\usepackage[T1]{fontenc}

\usepackage[utf8]{inputenc}

\usepackage{microtype}

\usepackage{inconsolata}

\usepackage{pifont}
\usepackage{caption}
\usepackage{adjustbox}
\usepackage{longtable}
\usepackage{array}
\usepackage[utf8]{inputenc}
\usepackage{booktabs}
\usepackage{graphicx} 
\usepackage{amsmath}
\usepackage{amssymb}

\definecolor{darkblue}{rgb}{0, 0, 0.5}
\hypersetup{colorlinks=true, citecolor=darkblue, linkcolor=darkblue, urlcolor=darkblue}

\title{Rethinking what Matters: Effective and Robust \\ Multilingual Realignment for Low-Resource Languages}

\author{Quang Phuoc Nguyen$^1$\thanks{Equal contribution.}, David Anugraha$^{2*}$, Felix Gaschi$^{3*}$, \\
\textbf{Jun Bin Cheng}$^{1}$, \textbf{En-Shiun Annie Lee}$^{1, 4}$ \\
  $^{1}$Ontario Tech University$\quad^{2}$Stanford University$\quad^{3}$SAS Posos$\quad^{4}$University of Toronto \\
\texttt{quangphuoc.nguyen@ontariotechu.net, david.anugraha@stanford.edu, felix@posos.fr}}

\newcommand{\xmark}{\mbox{\ding{55}}}

\begin{document}
\maketitle
\begin{abstract}

Realignment is a promising strategy to improve cross-lingual transfer in multilingual language models. However, empirical results are mixed and often unreliable, particularly for typologically distant or low-resource languages (LRLs) compared to English. Moreover, word realignment tools often rely on high-quality parallel data, which can be scarce or noisy for many LRLs. In this work, we conduct an extensive empirical study to investigate whether realignment truly benefits from using all available languages, or if strategically selected subsets can offer comparable or even improved cross-lingual transfer, and study the impact on LRLs. Our controlled experiments show that realignment can be particularly effective for LRLs and that using carefully selected, linguistically diverse subsets can match full multilingual alignment, and even outperform it for unseen LRLs. This indicates that effective realignment does not require exhaustive language coverage and can reduce data collection overhead, while remaining both efficient and robust when guided by informed language selection.\footnote{Our code can be found at~\url{https://github.com/felixgaschi/multilingual-alignment-and-transfer}.}
\end{abstract}

\section{Introduction}
\label{sec:intro}

Multilingual pre-trained language models such as mBERT~\citep{devlin2019bert} and XLM-R~\citep{conneau-etal-2020-unsupervised} enable cross-lingual transfer, where models fine-tuned to a certain task with an English dataset can be generalized to the same task in other languages~\citep{pires2019multilingual, wu2019beto}. However, their performance often degrades for typologically distant languages, such as low-resource languages (LRLs)~\citep{pires2019multilingual}. A promising strategy to address this issue is to perform realignment, which explicitly retrains models to produce similar representations for translated sentence pairs using objectives inspired by multilingual word embeddings~\citep{conneau2017word, artetxe2018robust}.

Despite a strong correlation between alignment and cross-lingual transfer~\citep{gaschi2023exploring}, results from realignment methods remain mixed. While some studies report benefits of realignment~\citep{cao2020multilingual, zhao2020inducing}, others observe limited or even negative effects~\citep{wu2020explicit, efimov2023impact}. These findings align with previous observations, where multilingual models exhibit good alignment for closely related languages, but remain more misaligned for distant or LRLs~\citep{dou2021word}.

In addition, realignment is not always feasible for all languages. It requires high-quality translation data, which may be unavailable for many LRLs~\citep{gu2018universal, liu2021continual, anugraha2024proxylm}. Even when resources exist, alignment quality can vary significantly across languages, potentially degrading downstream performance. This raises a central question: \textbf{Do we need to use all available languages for a better realignment, or could a carefully selected subset of languages offer similar or improved cross-lingual transfer performance?}

\begin{figure*}[!th]
    \centering
    \includegraphics[width=0.95\textwidth]{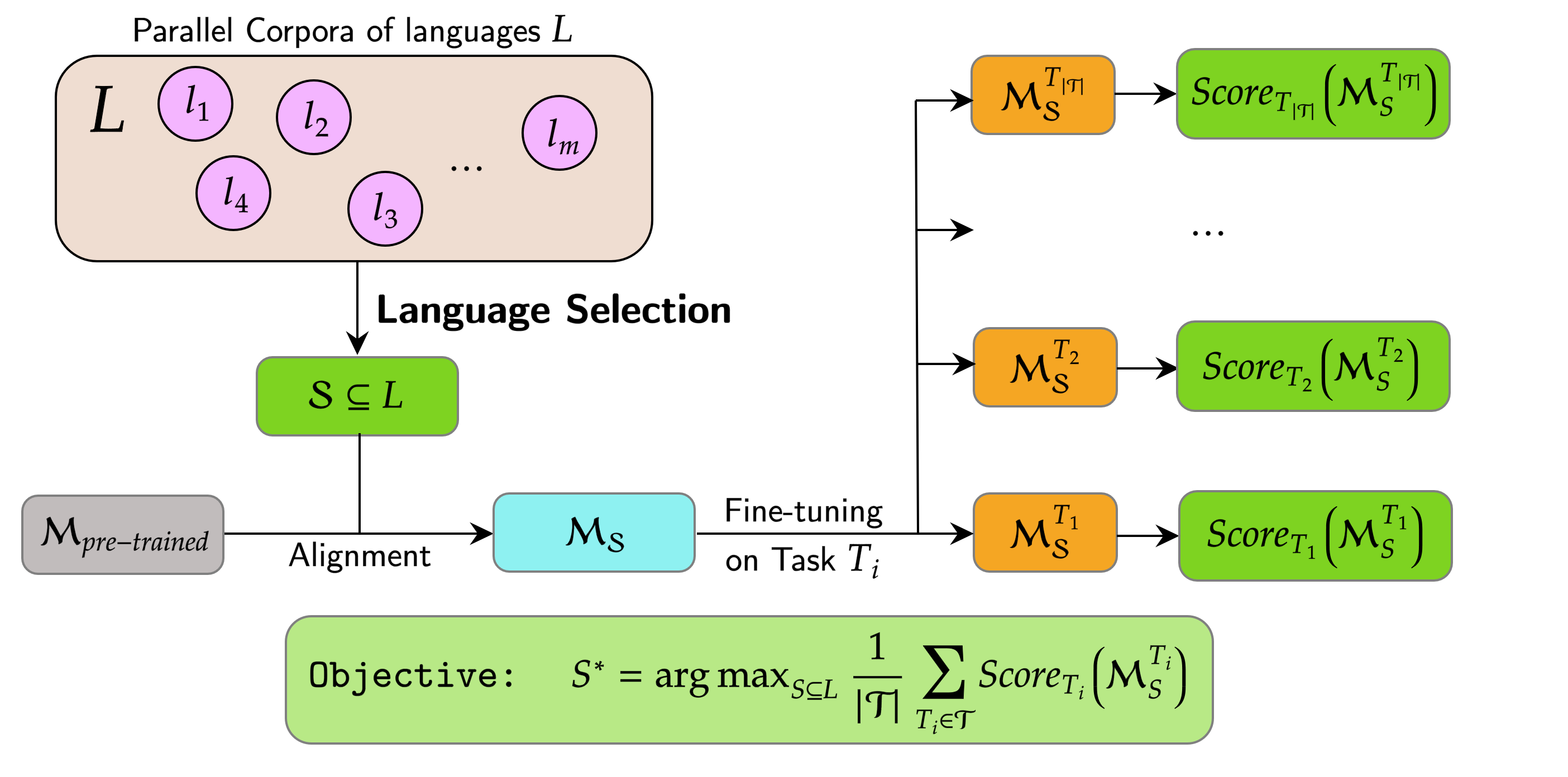}
    \caption{Overall diagram of the realignment process. Our goal is to empirically investigate how language selection within the realignment dataset impacts overall downstream task performance.}
    \label{fig:main_method}
\end{figure*}



In this work, we conduct an extensive empirical study to investigate whether realignment truly benefits from using all available languages, or if strategically selected subsets can offer comparable or even improved cross-lingual transfer. In summary, our key contributions are:

\begin{enumerate}
    \item \textbf{We conduct the first large-scale, systematic evaluation of realignment across  65 languages}, including 29 LRLs, 3 tasks, 4 seeds, and 2 models (with a strong focus on low-resource scenarios). By introducing a sentence-level averaging and contrastive objective that removes the need for word aligners, we show significant gains of up to 10 points in cross-lingual transfer, especially for LRLs unseen during pre-training.
    \item \textbf{We systematically investigate language subset selection for efficiency}, demonstrating that informed subsets chosen via heuristics (like URIEL featural diversity) can match or surpass full multilingual realignment. This shows that linguistic diversity matters more than the sheer number of languages.
    \item \textbf{We perform comprehensive ablation studies, including out-of-distribution robustness}. We evaluate on unseen, out-of-distribution benchmarks (e.g., AmericasNLI) to show that diverse subset selection generalizes effectively. We also conduct ablations by scaling the number of languages and varying initial language pools to reflect realistic resource constraints, showcasing the importance of including LRLs in realignment.
\end{enumerate}

To the best of our knowledge, we are the first to evaluate realignment massively on truly LRLs.

\section{Methodology}
\label{sec:method}

Recall that we perform realignment to explicitly retrain multilingual encoders to produce similar representations for translated sentence pairs. In particular, we first perform realignment as a separate training phase, which is then followed by full-model fine-tuning on a downstream task, following previous work~\citet{wu2020explicit,gaschi2023exploring,bakos-etal-2025-alignfreeze}. For the realignment phase, we adopt the method proposed by~\citet{wu2020explicit}, which modifies the encoder to produce similar representations for semantically equivalent words across languages. This is achieved using a contrastive loss applied to word-level alignment pairs extracted from parallel corpora.

Prior work has typically relied on extracting word pairs from parallel sentences using word aligners such as FastAlign~\citep{dyer-etal-2013-simple} or bilingual dictionaries~\citet{gaschi2023exploring}. However, these alignment resources are often unreliable or entirely unavailable, especially for LRLs, and their use typically requires substantial computational resources. Thus, we propose a simple alternative that removes the dependency on word aligners while requiring significantly less time and computational resources. Our method instead averages the representations of words in each sentence of a translation pair and directly minimizes the distance between these sentence-level representations.

Formally, let $B$ denote the batch size, and let $H = \{(h_i, \tilde{h}_i)\}_{i=1}^B$ represent a batch of $B$ aligned sentences, where $h_i$ is the averaged embedding of the words in a source (e.g., English) sentence and $\tilde{h}_i$ is the embedding of its aligned counterpart in the target language. The goal is to bring $h_i$ and $\tilde{h}_i$ closer together in the embedding space while pushing $h_i$ away from all other unaligned sentences in the batch. This is achieved via the following contrastive loss:

\begin{equation}
\mathcal{L}(\theta) = \frac{1}{2B} \sum_{h \in H} \log \frac{
\exp\left( \text{sim}(h, \text{aligned}(h)) / T \right)
}{
\sum\limits_{h' \in H,\; h' \neq h} \exp\left( \text{sim}(h, h') / T \right)
}
\end{equation} 

where $\text{sim}(h, h')$ denotes cosine similarity between two representations and $T$ is a temperature hyperparameter, set to 0.1 in our experiments.

Note that the contrastive loss defined above implicitly depends on the translation data used, particularly the set of languages involved. Prior work typically performs realignment using all available languages for their parallel data~\citep{wu2020explicit, gaschi2023exploring, bakos-etal-2025-alignfreeze}. In contrast, we hypothesize that a carefully selected \emph{subset} of languages may suffice to achieve comparable or even improved downstream cross-lingual generalization.

Formally, let $L = \{\ell_1, \ell_2, \dots, \ell_m\}$ be the full set of languages for which parallel corpora with English are available. Let $\mathcal{D}_S$ denote the parallel data involving English and the languages in subset $S \subseteq L$, and let $\mathcal{M}_S$ denote the model after realignment using this data. Let $\mathcal{T} = \{T_1, T_2, \dots, T_k\}$ be a set of downstream tasks, and for each task $T_i \in \mathcal{T}$, we fine-tune the realigned model $\mathcal{M}_S$ on task-specific supervision to obtain the fine-tuned model $\mathcal{M}^{T_{i}}_{S}$. We then compute the corresponding evaluation score $\text{Score}_{T_i}(\mathcal{M}^{T_i}_{S})$. Overall, our goal is to find the subset $S^* \subseteq L$ that maximizes the macro-average across all downstream tasks:\footnote{This performance evaluation setup follows prior work in multi-task multilingual learning, such as by XTREME-R~\citep{ruder-etal-2021-xtreme}.}

\begin{equation}
S^* = \mathop{\arg\max}_{S \subseteq L} \dfrac{1}{|\mathcal{T}|} \sum_{T_i \in \mathcal{T}} \text{Score}_{T_i}\left(\mathcal{M}^{T_i}_{S}\right)
\end{equation}

Since the downstream evaluation metric is non-differentiable and trying all possible subsets of $L$ is expensive, we do not optimize this objective directly. Instead, we construct subsets using linguistic-motivated heuristics as will be described in Section~\ref{sec:lang-subsets}.

\section{Experimental Setup}
\label{sec:experimental-setup}

\subsection{Language Subsets}
\label{sec:lang-subsets}

We construct and evaluate subsets of languages based on heuristics designed to capture different dimensions of cross-lingual diversity and coverage. These heuristics consider factors such as linguistic feature diversity, language family affiliation, and script variation. To assess the effectiveness of these heuristics, we also compare their performance against randomly selected realignment languages, thereby evaluating the significance of each heuristic.

All subsets are drawn from the same pool of 65 languages, which we denote as \(L_{65}\). This pool consists of 47 languages from XTREME-R~\citep{ruder-etal-2021-xtreme} together with 21 additional African languages, with some overlap between the two groups. The sets of 21, 47, and 65 languages serve as our baseline subsets. Details of the languages in \(L_{65}\) are provided in Table~\ref{tab:languages}. For each heuristic, we evaluate subsets of size \(n \in \{5, 10, 20, 40\}\), corresponding to increasing coverage when available.

\paragraph{Baselines.} We include two types of baselines to contextualize the performance of our subset selection strategies. The first baseline uses the fixed language sets of size 21, 47, and 65 as mentioned above. The second baseline consists of random subsets sampled uniformly from \(L_{65}\) with \(n \in \{5, 10, 20, 40\}\). These baselines help distinguish the effect of informed linguistic heuristics from arbitrary selection. All random subsets are generated using fixed random seeds for reproducibility.


\paragraph{Language Featural Diversity.} This heuristic aims to compute diversity for languages based on their structural linguistic features. These features are obtained using the URIEL+ database~\citep{khan-etal-2025-uriel}, which is a language vector resource that encodes languages based on typological, geographic, phonological, syntactic, and phonetic inventory feature vectors. These representations allow for the computation of pairwise distances between languages, using angular distance over their vectorized representations. We compare two types of subsets: (1) subsets where we have the most diverse set of languages from set $L_{65}$, and (2) subsets where we have the least diverse set of languages from set $L_{65}$. To construct our most diverse subsets, we select languages that maximize their pairwise featural distance from English, with English included in the subset calculation but not considered during realignment. We also constructed the least diverse subsets to contrast with the diverse case by minimizing the total pairwise featural distance. The formal definition of the objective can be found in Section~\ref{app:diversity-calculation}.

\paragraph{Language Family Diversity.} This heuristic investigates whether diversity in genetic lineage contributes to effective realignment. We compare two types of subsets: (1) subsets where each language comes from a distinct language family other than the Indo-European family, and (2) subsets that are restricted to a single language family, specifically, Indo-European languages, to contrast with the diverse case.

\paragraph{Script Diversity.} This heuristic investigates whether diversity in language scripts contributes to effective realignment. We compare three types of subsets: (1) subsets where each language is drawn from a distinct script other than Latin, (2) diverse subsets (as defined in Language Featural Diversity) but restricted to languages in \(L_{65}\) that use only the Latin script, and (3) least diverse subsets restricted to the Latin script, serving as a contrast to the diverse case.

All the language subsets and their languages are listed in the Appendix.

\begin{figure*}[!th]
    \centering
    \includegraphics[width=\textwidth]{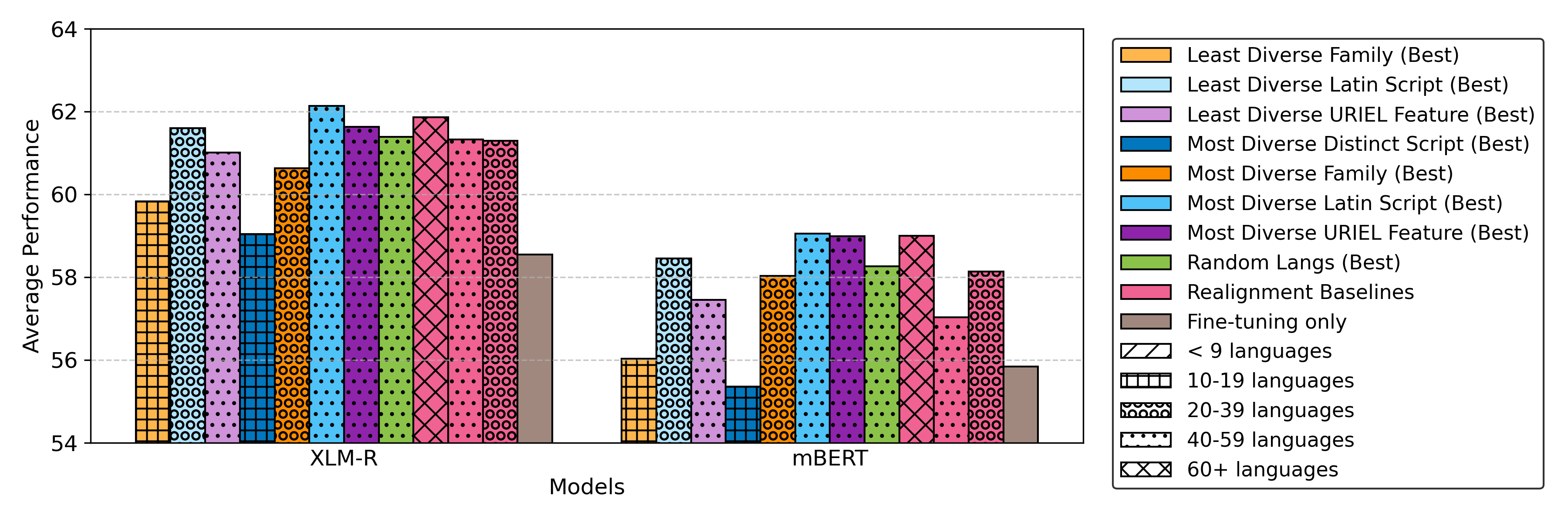}
    \caption{Average performance across PoS Tagging, NER, and NLI for XLM-R and mBERT. The baselines are compared against the best-performing configuration from each language subset heuristic.}
    \label{fig:main_results}
\end{figure*}

\subsection{Models and Datasets}

\paragraph{Realignment dataset} We use OPUS-100~\citep{zhang2020improving} and NLLB~\citep{costa2022no}, which contain parallel corpora with sentence pairs across 100 and 200 languages, respectively. Whenever a language is not covered by OPUS-100, we fall back to the NLLB dataset.

\paragraph{Training and Downstream Task Datasets} 
To evaluate cross-lingual transfer, we fine-tune all models exclusively on the English subset and evaluate them directly on other languages without additional fine-tuning.  Our evaluation is mostly focused on \emph{in-distribution} datasets, where evaluation languages are part of the realignment language set. However, we also included an \emph{out-of-distribution (OOD)} dataset scenarios, which contain languages not seen during pre-training or realignment. Detailed dataset statistics are provided in Table~\ref{tab:dataset-statistics}.

\paragraph{In-Distribution Datasets.} 
We consider three downstream tasks: Part-of-Speech (PoS) tagging, Named Entity Recognition (NER), and Natural Language Inference (NLI), evaluated on datasets covering both the XTREME-R language set and their African counterparts. For each task, we evaluate on datasets covering both the XTREME-R language set and African counterparts:
\begin{itemize}
    \item For PoS tagging, we fine-tune on the UDPOS dataset~\citep{de2021universal} and evaluate on both UDPOS and MasakhaPOS~\citep{dione-etal-2023-masakhapos}.  
    \item For NER, we fine-tune on the English subset of WikiANN~\citep{pan-etal-2017-cross} and evaluate on WikiANN and MasakhaNER~\citep{adelani-etal-2022-masakhaner}.  
    \item For NLI, we fine-tune on the English subset of XNLI~\citep{conneau2018xnli} and evaluate on four datasets: XNLI, IndoNLI~\citep{mahendra-etal-2021-indonli}, Myanmar-XNLI~\citep{htet2025myanmar}, and AfriXNLI~\citep{adelani2024irokobench}. For AfriXNLI, we restrict evaluation to its African languages.
\end{itemize}
    
\paragraph{Out-of-Distribution Datasets.} 
To study generalization beyond the languages present during pre-training or realignment,  we evaluate NLI performance on AmericasNLI~\citep{ebrahimi2021americasnli}, which covers 10 typologically diverse languages absent from both the pre-training and realignment language sets.  This dataset serves as a challenging out-of-distribution benchmark to assess zero-shot cross-lingual transfer.

\paragraph{Models} We use mBERT~\citep{devlin2019bert} and XLM-R~\citep{conneau-etal-2020-unsupervised} as our multilingual pre-trained language models. All experiments are run on 4 different seeds. More details about the hyperparameters can be found in Section~\ref{app:hyperparameters}.

\section{Results and Analysis}
\label{sec:results}

\subsection{Results Overview}


Figure~\ref{fig:main_results} presents a comparison of the best average performance across different tasks for both XLM-R and mBERT, evaluated under different language subset heuristics. Detailed per-task results are reported in Tables~\ref{tab:results-detailed-xlmr} and~\ref{tab:results-detailed-mbert}.



First, Figure~\ref{fig:main_results} shows that realignment provides significantly better overall results than the fine-tuning baseline. Except for two selection methods, realignment provides at minimum a one-point improvement, indicating the benefits of having to perform realignment.

However, performing realignment using all languages is not necessary to achieve comparable or even better performance. Figure~\ref{fig:main_results} shows that the realignment strategy based on URIEL featural diversity and URIEL featural diversity within languages with Latin script consistently yields the best results, achieving performance comparable to using the full set of 65 realignment languages. This demonstrates that carefully selecting a smaller but linguistically diverse subset of languages can be as effective as, or even better than, using all languages. These two heuristics also outperform other baselines, including random selection, XTREME-R only, and African-only subsets.

Our results further highlight, across all heuristics, the least diverse subsets consistently underperform compared to their more diverse counterparts. Language subsets selected to maximize diversity in featural space outperform those that minimize such diversity across both models. Likewise, selecting languages from distinct families offers clear benefits over limiting realignment to a single family. These results show that realignment benefits from the inclusion of languages that provide diverse linguistic signals, probably because such signals help to anchor multilingual representations more robustly.

Finally, diversity affects realignment performance in different ways depending on the dimension of diversity considered. For example, diversity based on genetic lineage does not yield strong results, while selecting languages with distinct scripts, excluding Latin, produces the worst performance. This suggests that script diversity can be beneficial, but the absence of Latin script hurts alignment performance, likely due to English being the pre-training language.




\begin{figure*}[!th]
    \centering
    \includegraphics[width=\textwidth]{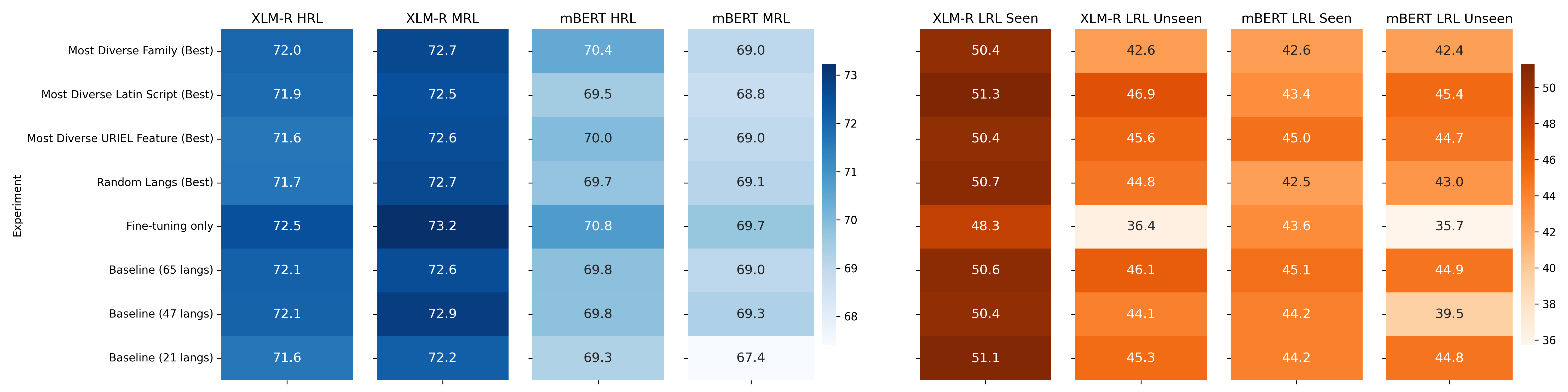}
    \caption{Heatmaps showing overall performance (averaged across four seeds) for different language subsets - HRLs, MRLs, and LRLs - seen and unseen during pre-training of XLM-R and mBERT. The fine-tuning only baseline remains strong for HRLs and MRLs, while realignment significantly improves performance on LRLs. Diversity-based language selection further amplifies these gains for LRLs.}
    \label{fig:heatmap-indistribution}
\end{figure*}

\subsection{Results Based on Language Resource Level}

To assess how different realignment methods impact different languages in terms of the level of resources, we categorize the evaluation languages into four groups: high-resource languages (HRLs), medium-resource languages (MRLs), low-resource languages (LRLs) that are seen during pre-training, and LRLs that are unseen during pre-training\footnote{HRLs = Joshi class 5, MRLs = 3 and 4, LRLs = 0, 1, and 2 \citep{joshi-etal-2020-state}}. Figure~\ref{fig:heatmap-indistribution} provides the detailed breakdown of overall performance for XLM-R and mBERT across different subsets of evaluation languages.

Realignment yields substantial gains for LRLs, particularly for languages unseen during pre-training. For both models, the best realignment configuration improves LRL-unseen performance by up to 10 points over standard fine-tuning. These results demonstrate that representation alignment is especially effective when cross-lingual transfer is weakest.

For HRLs and MRLs, the trends differ. Fine-tuning alone remains competitive on these languages, and applying realignment leads to slight performance drops. This pattern is consistent with prior findings that realignment benefits do not always extend to higher-resource languages~\citep{wu2020explicit,gaschi2023exploring}, which have stronger initial cross-lingual representations.

We also compare different strategies for selecting realignment languages. While language choice has little influence on HRLs or MRLs, it noticeably affects LRL performance. The advantage of diversity-based over random selection observed in aggregate results primarily stems from improvements on LRLs.

Overall, these results highlight a key insight: even though realignment may offer limited gains for HRLs and MRLs, it provides consistent and substantial improvements for LRLs, especially those absent from pre-training. This makes realignment a promising direction for extending multilingual encoders to truly underrepresented languages.

\subsection{Results on Out-of-Distribution Languages}

\begin{figure*}[!th]
    \centering
    \includegraphics[width=\textwidth]{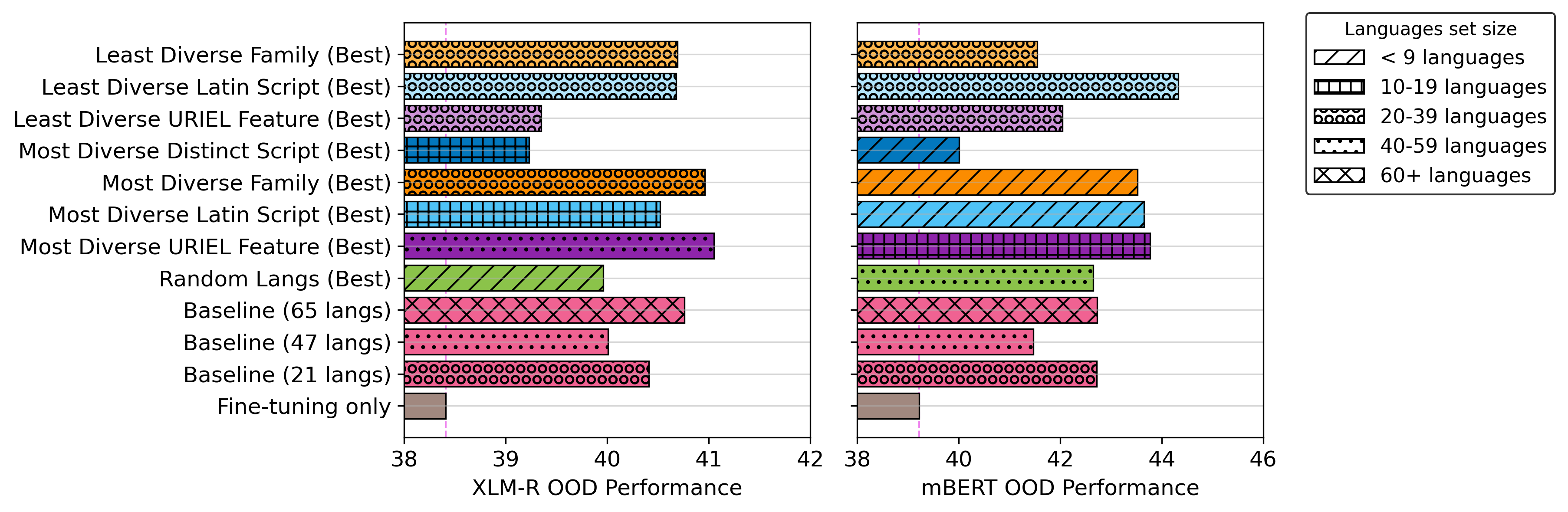}
    \caption{Averaged out-of-distribution performance of XLM-R and mBERT on the AmericasNLI dataset, comparing different language selection heuristics against three realignment baselines and a fine-tuning-only baseline. Realignment with diversity-based language subsets outperforms both the realignment and fine-tuning-only baselines.}
    \label{fig:americasNLI}
\end{figure*}

To complement our in-distribution analysis, we further evaluate different realignment approaches on AmericasNLI, which contains LRLs that were not used for realignment.

Figure~\ref{fig:americasNLI} shows that the results on LRLs unseen during realignment do not differ much from languages used for realignment. Similarly to Figure~\ref{fig:main_results}, realignment significantly outperforms the fine-tuning only baseline. Diversity-based language selection outperforms the random baseline, and their homegenous counterparts, with the exception of maximizing URIEL diversity within Latin-scripted languages, which suggests that diversity should be enforced in all aspects (featural, script, and family). 

One key difference with in-distribution results is that diversity-based selection, namely when using URIEL features, outperforms realignment on the entire set of available languages. Thus, when it comes to improving results across the board, including languages unseen during realignment, diversity might become more important than the number of languages involved.

As shown in Figure~\ref{fig:americasNLI}, realignment again substantially outperforms the fine-tuning baseline, mirroring the in-distribution trends. Between language subsets used for realignment, diversity-based selection continues to outperform both random selection and homogeneous subsets, with one exception: maximizing URIEL diversity within Latin-script languages does not provide the same advantage, suggesting that meaningful diversity must span features, scripts, and families rather than being constrained to a single script group. Furthermore, URIEL-based selection outperforms realignment on the full set of languages, demonstrating that when the goal is broad cross-lingual improvement, including languages never seen during realignment, the type of diversity in the realignment set matters more than the number of languages it contains.

\section{Language-Scaling Behavior of Realignment Methods}

\begin{figure*}[!th]
    \centering
    \includegraphics[width=\textwidth]{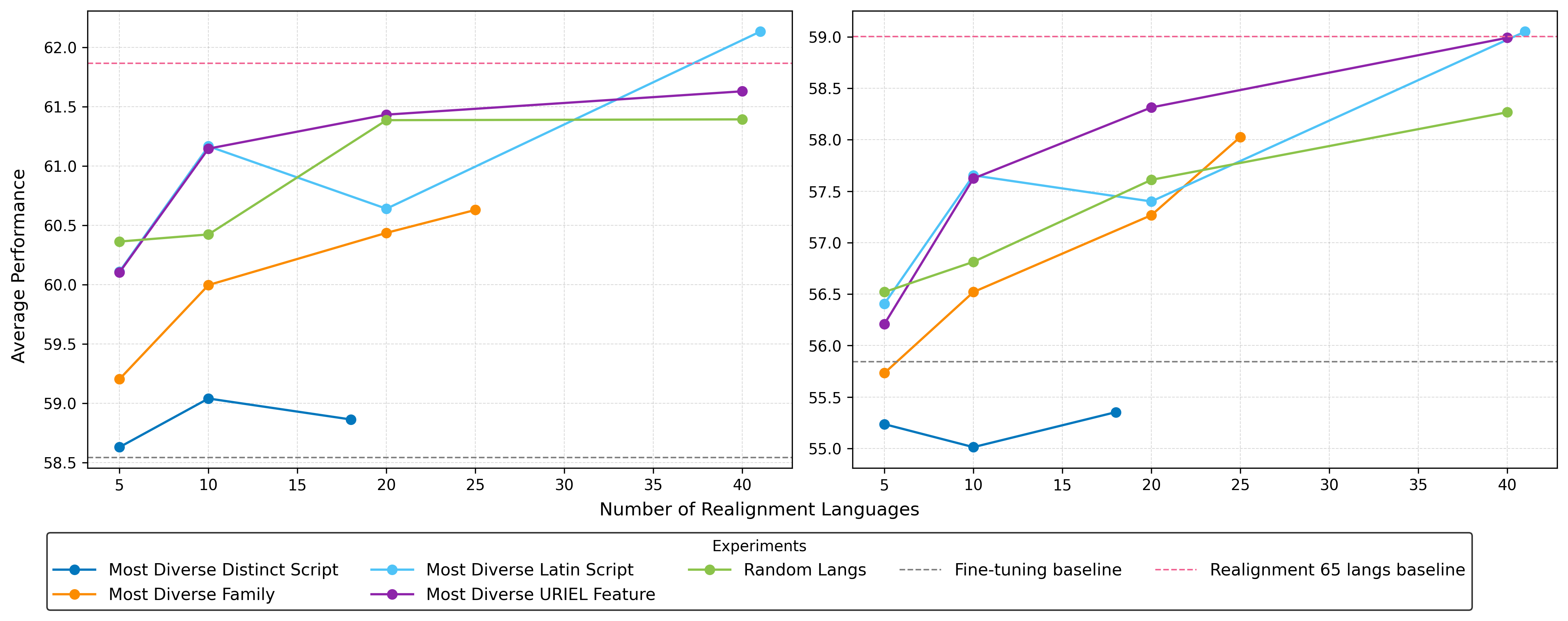}
    \caption{Scaling of average cross-lingual transfer performance with the number of languages used for realignment for XLM-R (left) and mBERT (right).}
    \label{fig:line-graph}
\end{figure*}

Figure \ref{fig:line-graph} shows how performance changes as we scale the number of languages used in each subset-selection strategy for realignment, while keeping the total computational budget fixed.

Across the board, every realignment strategy improves over simple fine-tuning, even with only five languages, indicating that cross-lingual realignment is beneficial even at very small scales. Among selection strategies, subsets based on distinct families or distinct scripts generally lag behind random sampling, whereas URIEL-diverse and diverse Latin-script language subsets provide stronger gains. Interestingly, the diverse Latin-script language subsets exhibit a non-monotonic trajectory, dipping from 10 to 20 languages before rising again at larger scales, suggesting that mid-scale expansions can occasionally introduce detrimental interactions before recovering.

For XLM-R, most strategies plateau around 20 languages, implying that the model absorbs most of the transferable signal once moderate coverage is reached. The diverse Latin-script language selection strategy is an exception, since the performance increases again at 40 languages and reversing its earlier dip. This suggests that additional gains still exist at large scales for strategies other than Latin-script diversity, but other optimal subsets may exist given a different language selection strategy.

For mBERT, the scaling behavior is more gradual and nearly linear. Several strategies, such as URIEL-based and diverse Latin-script language selection, even surpass the full 65-language baseline at intermediate scales. This indicates that mBERT continues to benefit from expanded cross-lingual supervision over a wider range than XLM-R, and that its saturation point occurs later.

Overall, realignment is consistently beneficial, and that URIEL-based diversity and diverse Latin-script language selection are the most reliable and data-efficient approaches across different number of languages. We also find that different models follow distinct scaling dynamics: XLM-R saturates early, whereas mBERT accumulates gains more steadily across larger language sets.

\section{Ablation study}

\begin{table}[h!]
\centering
\adjustbox{max width=\linewidth}{
\begin{tabular}{lcccc}
\toprule
\textbf{Language Pool} & \textbf{POS} & \textbf{NLI} & \textbf{NER} & \textbf{Avg.}\\
\midrule
\multicolumn{5}{l}{\textbf{XLM-R}} \\
\addlinespace
Joshi 4 and 5 & 67.5 & 58.6 & 54.4 & 60.2\\
Joshi 3, 4, and 5 & 67.2 & 58.8 & 53.8 & 59.9\\
Joshi 3 & 67.0 & 58.9 & 51.0 & 59.0\\
Joshi 2 & 68.8 & \textbf{59.9} & \textbf{54.5} & \textbf{61.1}\\
Unseen Languages & \textbf{69.1} & 59.8 & 53.5 & 60.8\\
Seen Languages & 67.1 & 58.8 & 53.3 & 59.7\\
\midrule
\multicolumn{1}{l}{\textit{Fine-tuning only}} & 66.0 & 58.6 & 51.1 & 58.6\\
\multicolumn{1}{l}{\textit{65 langs baseline}} & 69.1 & 59.4 & 57.1 & 61.9\\
\midrule
\multicolumn{5}{l}{\textbf{mBERT}} \\
\addlinespace
Joshi 4 and 5 & 63.7 & 53.3 & 50.7 & 55.9\\
Joshi 3, 4, and 5 & 63.5 & 53.3 & 51.5 & 56.1\\
Joshi 3 & 62.9 & 53.2 & 50.3 & 55.5\\
Joshi 2 & \textbf{65.1} & \textbf{55.0} & \textbf{52.5} & \textbf{57.5}\\
Unseen Languages & 64.5 & 54.4 & \textbf{52.5} & 57.1\\
Seen Languages & 63.7 & 53.4 & 51.2 & 56.1\\
\midrule
\multicolumn{1}{l}{\textit{Fine-tuning only}} & 62.2 & 53.1 & 52.2 & 55.8\\
\multicolumn{1}{l}{\textit{65 langs baseline}} & 66.9 & 55.4 & 54.8 & 59.0\\
\bottomrule
\end{tabular}
}
\caption{Performance of different realignment strategies for XLM-R and mBERT under a 10-language constraint. Only the random language selection strategy is shown. Bold indicates the highest result per task and model (excluding baselines). Standard deviation and results for other selection strategies are shown in Tables \ref{tab:results-ablation-xlmr} and \ref{tab:results-ablation-mbert} in Appendix.}
\label{tab:ablation_table}
\end{table}

In this ablation study, we focus specifically on analyzing the impact of including languages of different resource levels in the realignment mix. Specifically, we consider the case where only limited resources are available to collect high-quality parallel data for realignment. Our goal is to to determine whether, under such constraints, lower-resource or unseen languages remain necessary for improving cross-lingual transfer, or if higher-resource languages can serve as effective substitutes. For the sake of clarity, rather than reporting results for all heuristics, we include only the random selection heuristic, alongside two baselines: fine-tuning only, and realignment using the entire $L_{65}$ set. Random language selection helps isolate the effect of different language pools on realignment performance, which is the main objective of this ablation study. Results for other selection heuristics can be found in Tables \ref{tab:results-ablation-xlmr} and \ref{tab:results-ablation-mbert} in the Appendix.



We compare performance when randomly sampling 10 languages from different pools for realignment: higher-resource languages only (Joshi 4 and 5), MRLs only (Joshi 3), a mixed set of MRLs and HRLs (Joshi 3, 4 and 5), and languages either seen or unseen during pre-training. Our results from Table~\ref{tab:ablation_table} show that in most cases, performing realignment on only 10 languages leads to improved cross-lingual transfer performance across tasks compared to fine-tuning alone. While using a reduced set of 10 languages for realignment does result in a performance drop relative to the 65-language baseline, the decrease is modest, ranging from 0.8\% to 1.5\%. This is a reasonable trade-off given the over sixfold reduction in the number of languages involved. On the other hand, comparisons among the ablation experiments reveal that language pools composed of Joshi Class 2 languages and unseen pretraining languages tend to yield better performance than other configurations. This highlights the importance of including LRLs and unseen languages in the realignment process to improve transfer on these same categories. 


Our ablation results also indicate that other language pools - such as mid- to high-resource languages from Joshi Classes 4–5 in the case of XLM-R, as well as seen pretraining languages - can serve as practical substitutes for LRLs when the latter are not applicable. Although there is a performance drop, it remains relatively minor (less than 1\%), while the availability of high-quality parallel data from these higher-resource language pools is considerably more likely. 

\section{Related Work}
\label{sec:related-work}

\paragraph{Realignment Strategies} Realignment typically involves two components: the \emph{alignment tool}, which identifies word correspondences between languages, and the \emph{training strategy}, which updates model parameters to enforce alignment \citep{hammerl-etal-2024-understanding}. Different alignment tools can be used, such as the statistical FastAlign~\citep{dyer-etal-2013-simple} and the neural AwesomeAlign~\citep{dou2021word}. For our training strategy, we adopt the method proposed by \citet{wu2020explicit}, which performs realignment to all layers by using a contrastive loss to word-level alignment pairs extracted from parallel corpora. Alternative training strategies include contrastive frameworks with different loss formulations \citep{chen2020simpleframeworkcontrastivelearning}, or architectural choices such as selectively realigning specific model layers \citep{bakos-etal-2025-alignfreeze}.

\paragraph{Data Selection Strategies} Multiple works across machine learning research have been able to show that strategic data selection, rather than using all available data, can lead to better generalization, efficiency, and robustness~\citep{wang2019target, albalak2024survey, liu2024regmix, anugraha2025r3}. In the context of cross-lingual transfer, prior work has shown that linguistic similarity between source and target languages is a strong predictor of transfer success, often outperforming naive or pivot-based strategies~\citep{duong-etal-2015-cross, Eronen_2023}. However, the optimal set of source languages depends heavily on the task, due to divergences in features like morphology, syntax, and script~\citep{philippy-etal-2023-towards}. In contrast to methods like LangRank~\citep{lin19acl}, which identify the best single transfer language per target language and downstream task, our work seeks to identify combinations of languages that optimize the average downstream performance across multiple target languages.

\section{Conclusion}
\label{sec:conclusion}


In this paper, we investigate whether realigning multilingual models with carefully selected language subsets can match or even surpass alignment using the full language set using linguistic-motivated heuristics. Our large-scale experiments demonstrated that realignment with a smaller language subsets often match the full set across models and tasks, especially when chosen based on their linguistic diversity, and evaluated on out-of-distribution languages. Moreover, our analysis shows that realignment most benefits LRLs, suggesting that realignment is particularly effective for languages whose embeddings are not yet well aligned. Our ablation studies further reveal that when the number of languages is limited, having LRLs in the language subset yields the strongest improvements, although HRLs and MRLs can still help enhance cross-lingual transfer. These results demonstrate that strategic language selection not only reduces computational and data overhead but can also strengthen multilingual generalization, pointing toward more efficient and targeted approaches to cross-lingual realignment.


\section*{Limitations}

In this paper, we demonstrate the importance of linguistic diversity as a more inclusive and effective approach to improving cross-lingual generalization in multilingual language models. Rather than simply collecting all available data, our work shows that carefully selecting diverse subsets of languages can enhance cross-lingual transfer, indicating that our work is both more efficient in terms of data collection overhead and more effective overall.

A limitation of this study is the absence decoder-only models, which are increasingly prevalent in current research. While the Appendix Section \ref{decoder_only_realignment_results} presents our tentative realignment results on Llama 3.1 8B using LoRA adapters, these results indicate that the same realignment method does not straightforwardly transfer to decoder-only architectures, highlighting the need for careful adaptation in such settings. Moreover, encoder-only architectures remain highly relevant for cross-lingual classification due to their efficiency, stable representations, and strong transfer capabilities. Recent developments, such as ModernBERT~\cite{warner-etal-2025-smarter} and LLM2Vec~\cite{behnamghader2024llm2veclargelanguagemodels}, which adapt decoder-only models into encoder-style architectures, further highlight their enduring importance. As shown in our small-scale experiment in Section~\ref{encoder_vs_decoder}, encoder-based models can even outperform much larger multilingual decoder-only models on several classification tasks. 
Future work could also explore applying our algorithm-agnostic heuristics to decoder-only model fine-tuning for multilingual tasks~\cite{anugraha2025mr3} or to reinforcement learning setups in multilingual contexts~\cite{dang2024rlhf}.

Another is that although averaging does not explicitly target word-level alignment, we empirically find that it provides results that are slightly lower but still comparable to FastAlign (Appendix \ref{avg_vs_fastalign}). Therefore, we opt for the averaging approach for practical reasons, as FastAlign is significantly more resource- and time-intensive.

Our exploration is also limited to heuristic-based language selection, although we have shown that such subsets do exist. A promising direction for future work is to move beyond heuristics by developing predictive algorithms that estimate downstream performance and dynamically determine both the optimal languages and the appropriate subset size \cite{anugraha2024proxylm}.

We acknowledge that our full realignment language set, $L_{65}$, does not cover the entire spectrum of global linguistic diversity, despite covering many LRLs. We hope that our approach encourages the creation of parallel corpora for underrepresented languages, enabling greater diversity in alignment sets and fostering more inclusive multilingual models. Ultimately, we aim for our work to contribute toward broader and fairer access to language technologies, especially in the context of cross-lingual NLP research and deployment.

\section*{Acknowledgements}
This research on "Multilingual multicultural NLP and LLMs" was supported by the Natural Sciences and Engineering Research Council of Canada (NSERC) Discovery Grant (RGPIN-2024-06887) and Discovery Launch Supplement (DGECR-2024-00008). The authors also acknowledge the computational resources and support provided by the Digital Research Alliance of Canada (formerly Compute Canada) through grant RRG no. 5397.

This project was provided with HPC computing and storage resources by GENCI at IDRIS thanks to the grant 2025-AD010316268 on the supercomputer Jean Zay's H100 partition.

\bibliography{custom}

\newpage

\appendix

\section{Detailed Methodology and Experimental Setup}

\subsection{Language Featural Diversity Calculation}
\label{app:diversity-calculation}

Let $L = \{\ell_1, \ell_2, \dots, \ell_m\}$ be the list of all available languages, $\text{vec}(\ell)$ denote the URIEL featural vector of language $\ell$, and let $d(u, v)$ denote the angular distance between vectors $u, v \in \mathbb{R}^d$. The angular distance is defined as:
\[
d(u, v) = \frac{\arccos\left( \frac{u \cdot v}{\|u\| \, \|v\|} \right)}{\pi}
\]
where:
\begin{itemize}
    \item $u \cdot v$ is the dot product of $u$ and $v$,
    \item $\|u\|$ is the Euclidean norm of vector $u$,
    \item $d(u, v) \in [0, 1]$ (normalized angle between vectors).
\end{itemize}

In order to find the most diverse subset size $n$, we try to maximize the total pairwise angular distance:
\[
S^* = \underset{S \subseteq L,\, |S| = n}{\arg\max} \;
\sum_{\{\ell_i, \ell_j\} \in \binom{S}{2}} d\left( \text{vec}(\ell_i), \text{vec}(\ell_j) \right)
\]

Similarly, to find the least diverse subset size $n$, we try to minimize the total pairwise angular distance:
\[
S^* = \underset{S \subseteq L,\, |S| = n}{\arg\min} \;
\sum_{\{\ell_i, \ell_j\} \in \binom{S}{2}} d\left( \text{vec}(\ell_i), \text{vec}(\ell_j) \right)
\]

\subsection{Realignment using average words' representations  vs FastAlign}
\label{avg_vs_fastalign}

\begin{table}[h]
\centering
\resizebox{\columnwidth}{!}{
\begin{tabular}{llcc}
\toprule
\textbf{Model} & \textbf{Task} & \textbf{Avg. Tokens} & \textbf{FastAlign} \\
\midrule
\multirow{3}{*}{mBERT}
    & NER & 54.89 {\scriptsize $\pm$ 0.93} & \textbf{55.39 {\scriptsize $\pm$ 0.52} }\\
    & NLI & 55.34 {\scriptsize $\pm$ 0.28} & \textbf{56.58 {\scriptsize $\pm$ 0.25} }\\
    & POS & 66.85 {\scriptsize $\pm$ 0.24} & \textbf{70.31 {\scriptsize $\pm$ 0.15} }\\
\midrule
\multirow{3}{*}{XLM-R}
    & NER & \textbf{57.19 {\scriptsize $\pm$ 0.79}} & 56.72 {\scriptsize $\pm$ 1.62} \\
    & NLI & 59.5 {\scriptsize $\pm$ 0.27} & \textbf{60.65 {\scriptsize $\pm$ 0.33}} \\
    & POS & 69.1 {\scriptsize $\pm$ 0.22} & \textbf{71.29 {\scriptsize $\pm$ 0.25}} \\
\cmidrule(lr){2-4}
\multicolumn{2}{l}{\textbf{Avg. Time (h)}} 
    & \textbf{0.35 {\scriptsize $\pm$ 0.003}} 
    & 1.93 {\scriptsize $\pm$ 0.039} \\
\bottomrule
\end{tabular}
}
\caption{Comparison of Average and FastAlign performance across 5 random seeds. Avg. Time reflects the average time taken by realignment step only across all tasks (in hours); each seed reused the same realignment checkpoint.}
\label{tab:fastalign_comparison}
\end{table}

Realignment methods introduced before this work take a batch of pairs of translated sentences, extract pairs of corresponding words across those pairs using alignment tools like FastAlign and AwesomeAlign, and compute an in-batch contrastive loss on those pairs of words. Our work introduces a simple “averaging trick”: instead of computing the contrastive loss on word pairs extracted with an aligner, we compute the average of all tokens in a sentence and align sentences instead of words. This change is not made to improve cross-lingual transfer but rather to alleviate the need for a word aligner, which considerably reduces the time necessary to perform realignment.

We compare the time efficiency and performance of the two realignment methods using a different GPU type from that used in our main experiments, with the results presented in Table~\ref{tab:fastalign_comparison}. Overall, FastAlign performs slightly better than the token averaging method on most tasks. However, we adopt the averaging approach for practical reasons, as FastAlign is considerably more resource- and time-intensive - even without accounting for the additional preprocessing required to prepare datasets for FastAlign across 65 languages - making it unsuitable for large-scale experiments. It is important to emphasize that we employ the averaging method \textbf{not to enhance the baseline performance}, but rather \textbf{to reduce computational overhead}, thereby enabling large-scale comparisons across different language selection strategies.

\subsection{Tentative Decoder-only Results}
\label{decoder_only_realignment_results}

\begin{table}[h]
\caption{Comparison of Llama model performance with and without realignment on downstream tasks.}
\label{tab:llama-realignment}
\centering 
\begin{tabular}{lcc}
\hline
\textbf{Metric} & \textbf{Llama w/o} & \textbf{Llama w/} \\
& \textbf{realignment} & \textbf{realignment}\\ 
\hline
PoS target acc. & 38.8 & \textbf{41.9} \\
NER target F1 & \textbf{31.6} & 30.2 \\
NLI accuracy & 56.5 & \textbf{57.8} \\
\hline
\end{tabular}
\end{table}

We experimented with Llama 3.1 8B using LoRA adapters to overcome computational limitations. The adapters were trained for realignment and then fine-tuned for the downstream task, following a similar procedure as for encoder-only models. The results are shown in Table \ref{tab:llama-realignment}.

While realignment improved PoS tagging and NLI, we observed that it negatively impacted NER performance for LLaMA, which suggests that \textbf{realignment does not transfer straightforwardly to decoder-only architectures.}. Our preliminary experiments are not conclusive on whether realignment could ultimately benefit decoder-only models. We plan to investigate this further in future work.

More broadly, applying realignment to decoder-only LLMs for generative tasks raises unique challenges. In particular, \textbf{making such models entirely language-agnostic could exacerbate issues such as language confusion} \cite{marchisio-etal-2024-understanding}, where the model generates text in an unintended language. This highlights an important avenue for future work and motivates careful consideration of how realignment should be adapted for decoder-only settings.

\subsection{Encoder-only models vs Decoder-only models on cross-lingual transfer classification}
\label{encoder_vs_decoder}

\begin{table}[htbp]
\centering
\begin{tabular}{lccc}
\toprule
\textbf{Task} & \textbf{XLM-R} & \textbf{Llama 3.1} & \textbf{Gemma 2} \\
\midrule
POS en & \textbf{96.2} & 90.9 & 93.5 \\
POS XL & \textbf{62.0} & 38.8 & 49.1 \\
\midrule
NER en & \textbf{82.0} & 72.5 & 73.9 \\
NER XL & \textbf{49.1} & 31.6 & 35.9 \\ 
\midrule
NLI en & 83.5 & 90.2 & \textbf{91.3} \\
NLI XL & 54.6 & \textbf{56.5} & 55.0 \\ 
\bottomrule
\end{tabular}
\caption{Performance comparison of XLM-R, Llama 3.1 8B, and Gemma 2 9B across various three downstream tasks: POS Tagging, NER and NLI under the same experimental settings. \textbf{XL} indicates cross-lingual.} 
\label{tab:model-comparison}
\end{table}

We fine-tuned LLaMA 3.1 8B on PoS, NER, and NLI under the same setup. Our results on Table \ref{tab:model-comparison} show that XLM-R (encoder-only) not only significantly outperforms Llama and Gemma on cross-lingual transfer for some tasks, but can even surpass them in English. Interestingly, this seems to be true for word-level tasks (POS and NER), but not sentence-level ones (NLI). This finding aligns with prior work suggesting that small fine-tuned encoder-only models often outperform prompted LLMs on classification tasks \cite{ahuja-etal-2023-mega}.

In conclusion, adapting realignment methods to encoder-only models and generative tasks is a modern and exciting direction for future research. Nevertheless, we focus on encoder-only models, which remain a relevant contribution for cross-lingual classification, especially for multilinguality.

\subsection{Languages}

Table~\ref{tab:languages} contains the full list of the 65 languages used in~\citet{gaschi2023exploring}. Table~\ref{tab:experiment-languages} contains the list of languages used in each experiment.

Table~\ref{tab:experiment-languages} shows the languages chosen within each experiment. All experiments are run with seeds of 17, 23, 42, and 66, including the selection of languages in the Random Subsets (Seeded) setting.

\begin{figure*}[!th]
    \centering
    \includegraphics[width=0.95\textwidth]{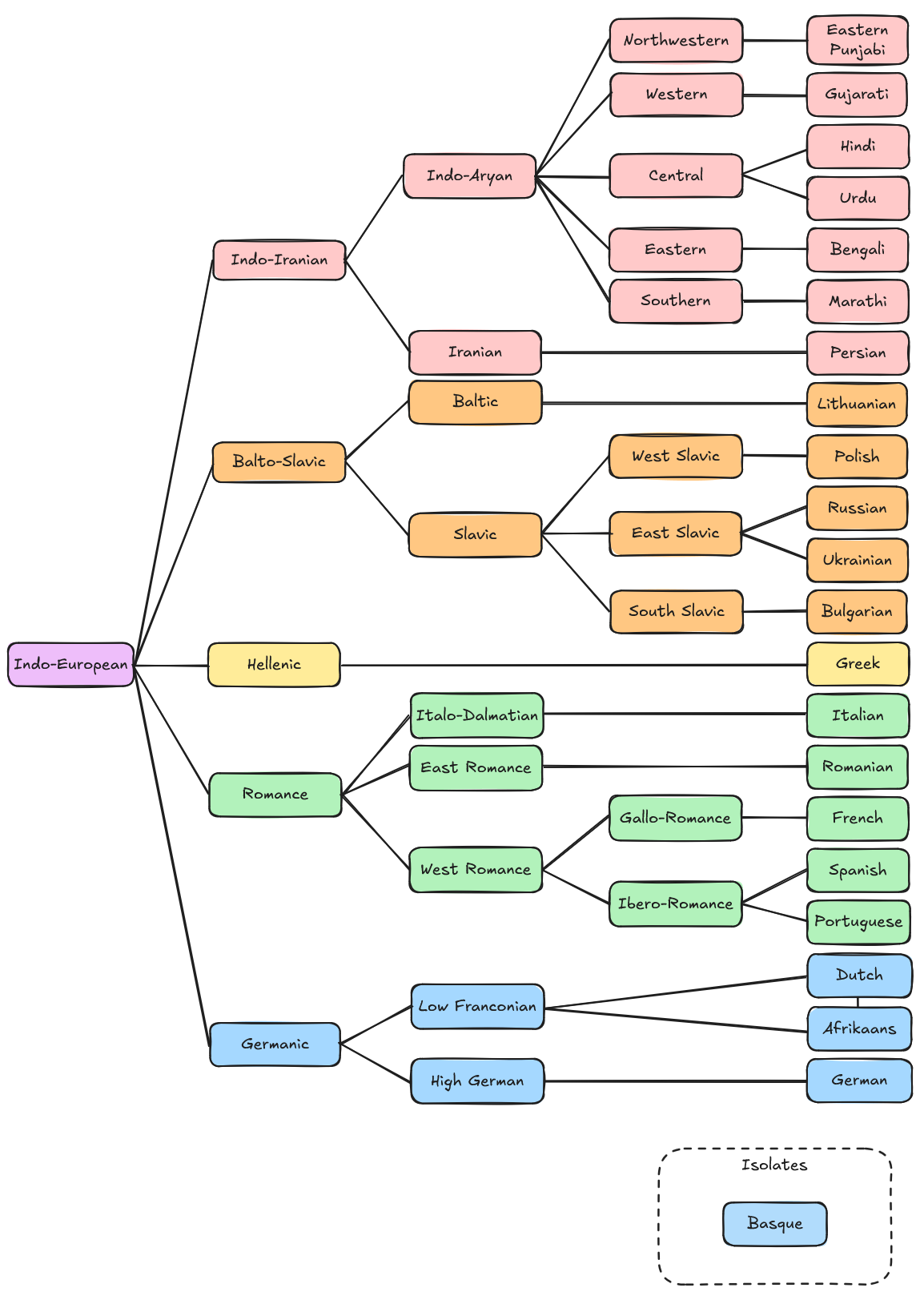}
    \caption{Phylogenetic tree for Indo-European languages (including Basque as an isolate) used in realignment.}
    \label{fig:indo-lang-tree}
\end{figure*}

\begin{figure*}[!th]
    \centering
    \includegraphics[width=0.95\textwidth]{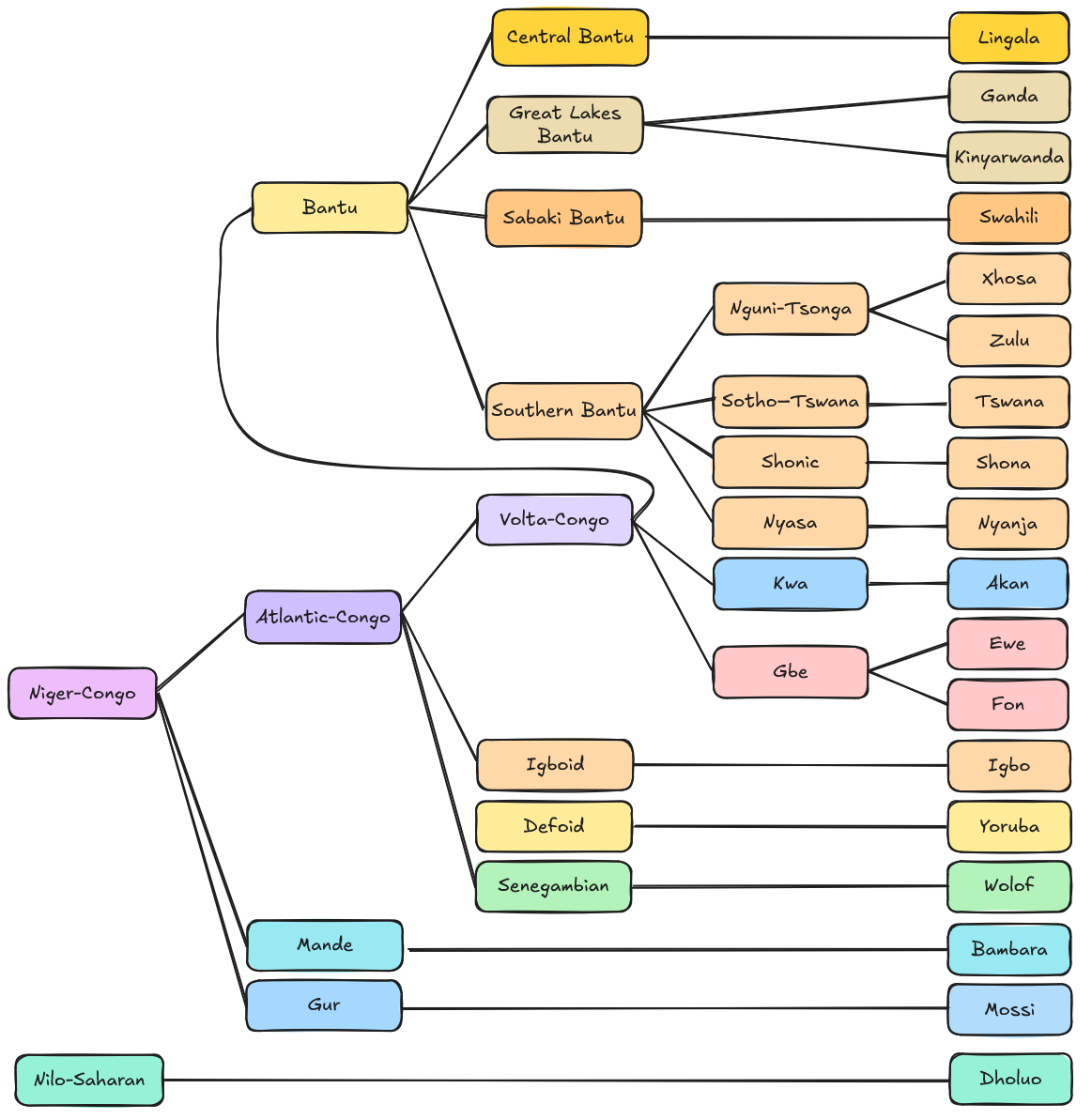}
    \caption{Phylogenetic trees for Niger-Congo languages, and Dholuo (from the Nilo-Saharan family, due to its proximity) used in realignment.}
    \label{fig:afri-lang-tree}
\end{figure*}

\begin{figure*}[!th]
    \centering
    \includegraphics[width=0.95\textwidth]{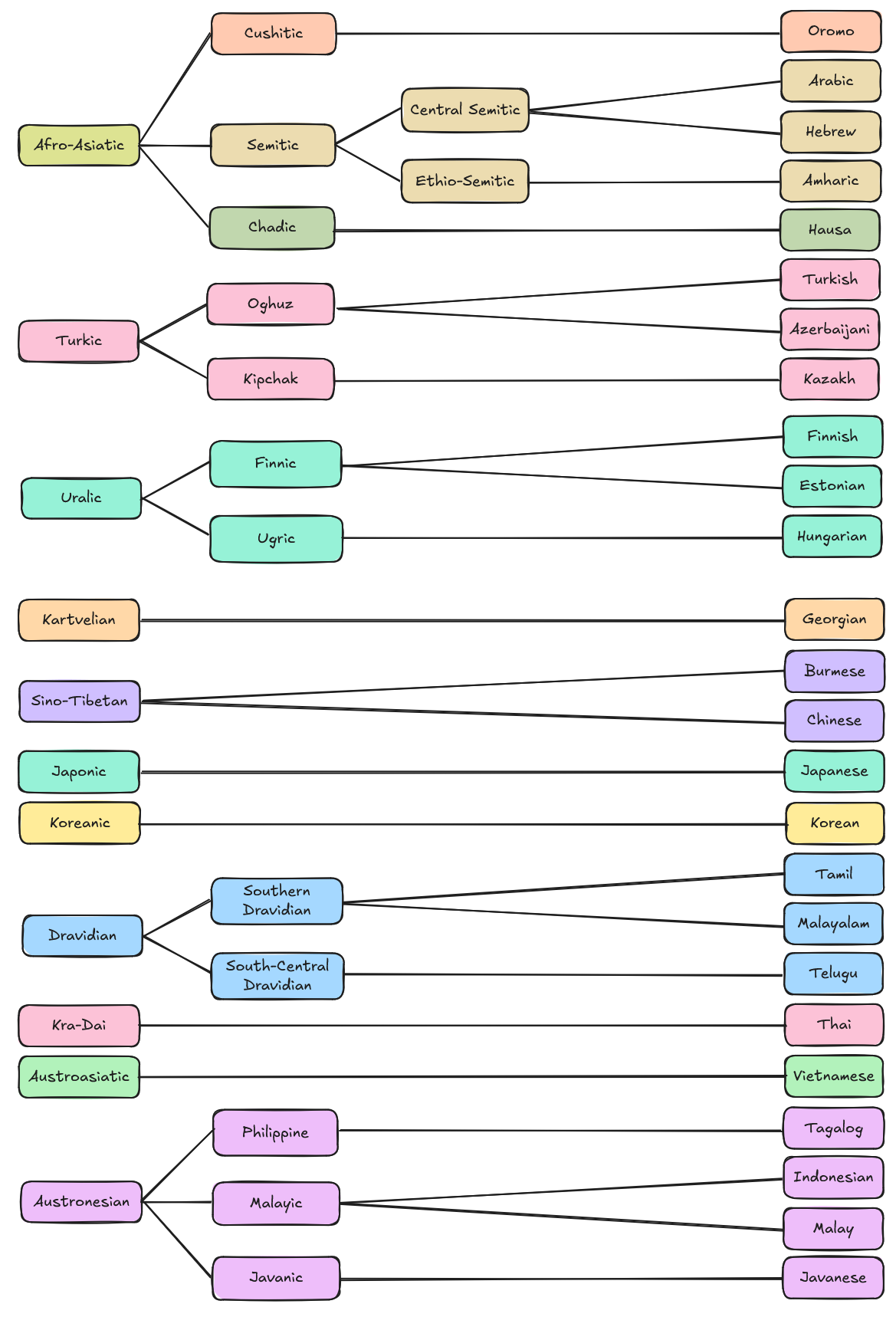}
    \caption{Residual phylogenetic trees for all other languages languages used in realignment (Non-Indo-European, Non-Niger-Congo, Non-Nilo-Saharan, Non-Basque).}
    \label{fig:other-lang-tree}
\end{figure*}

Figures~\ref{fig:indo-lang-tree}, ~\ref{fig:afri-lang-tree}, and  ~\ref{fig:other-lang-tree} show the language trees of the 65 languages used.

\subsection{Hyperparameters and Resources}
\label{app:hyperparameters}

For both tasks, we follow the experimental setup used in~\citet{gaschi2023exploring, bakos-etal-2025-alignfreeze}. All experiments are conducted using NVIDIA H100 GPUs and run with 4 random seeds to account for variability. Realignment is performed for 24,544 steps. This is followed by task-specific fine-tuning: 5 epochs for PoS tagging and 2 epochs for NLI. We use a learning rate of \texttt{2e-5}, a batch size of 32 for both training and evaluation, and a maximum input length of 200 tokens for source and target sequences. During the realignment stage, we use a reduced maximum sequence length of 96 and a smaller batch size of 16.

\subsection{Statistics about the datasets used}
\label{sec:datasets}
The size of the datasets used for training and evaluating are reported in Table \ref{tab:dataset-statistics}.

\subsection{Licenses for artifacts used}
\label{sec:licenses}
Below is a list of the datasets under study:
\begin{itemize}
    \item The XNLI corpus~\citep{conneau2018xnli} has the CC BY-NC 4.0 license.
    \item The AfriXNLI dataset~\citep{adelani2024irokobench} has the Apache 2.0 license.
    \item The IndoNLI dataset~\citep{mahendra-etal-2021-indonli} has the CC-BY-SA 4.0 license.
    \item The Myanmar-XNLI dataset~\citep{htet2025myanmarxnlibuildingdataset} has the Apache 2.0 license.
    \item The UDPOS dataset~\citep{de2021universal} has the CC0-1.0 license.
    \item The MasakhaPOS dataset~\citep{dione-etal-2023-masakhapos} has the MIT license.
    \item The WikiANN dataset~\citep{pan-etal-2017-cross} has the Apache 2.0 2.0 license.
    \item The MasakhaNER 2.0 dataset~\citep{adelani-etal-2022-masakhaner} has the AFL 3.0 license.
    \item The OPUS-100 dataset~\citep{zhang2020improving} has no explicit license; it is a filtered subset of OPUS~\citep{tiedemann2009news}, which aggregates translation corpora that is generally considered redistributable.  
    \item The NLLB dataset~\citep{nllbteam2022languageleftbehindscaling} has the ODC-By license.
    \item The XTREME-R benchmark suite~\citep{ruder-etal-2021-xtreme} does not have a unified license; it aggregates multiple datasets, each with its own license or terms of use:\\
    \begin{itemize}
        \item The XNLI corpus~\citep{conneau2018xnli} has the CC BY-NC 4.0 license.
        \item The PAWS-X dataset~\citep{Yang2019paws-x} is free to use for any purpose.
        \item The UDPOS dataset~\citep{de2021universal} has the CC0-1.0 license.
        \item The WikiANN dataset~\citep{pan-etal-2017-cross} has the Apache 2.0 2.0 license.
        \item The XQuAD dataset~\citep{artetxe2020cross} has the CC BY-SA 4.0 license.
        \item The MLQA dataset~\citep{Lewis2020mlqa} has the CC BY-SA 3.0 license.
        \item The TyDiQA-GoldP dataset~\citep{Clark2020tydiqa} has the Apache 2.0 license.
        \item The BUCC 2018 dataset for the shared task on bitext mining~\citep{zweigenbaum2018overview} is available for academic research use only; redistribution may be restricted.
        \item The Tatoeba dataset~\citep{Artetxe2019massively} has the CC BY 2.0 license.
    \end{itemize}
\end{itemize}

Below is a list of the other artifacts under study:
\begin{itemize}
    \item The code for realignment comes from \citet{gaschi2023exploring} and has the MIT license.
    \item The URIEL+ knowledge base and distance calculation functions \citep{khan-etal-2025-uriel} have the CC BY-SA 4.0 license.
    \item The weights of XLM-R Base ~\citep{conneau-etal-2020-unsupervised} have the MIT license.
    \item The weights of mBERT Base ~\citep{devlin2019bert} have the Apache 2.0 license.
\end{itemize}
All artifacts were thus used in accordance with their open-source or non-commercial licenses.

\subsection{Use of AI}

For the writing of this paper, AI was solely used to reformulate some text, and as an autocompletion tool for writing the code used in the experiments.

\section{More Detailed Results}

This section contains the full results of the experiments in this paper.

\captionsetup[longtable]{skip=10pt}
\onecolumn
    \begin{longtable}{@{}lccccc@{}}
        \toprule
            \textbf{Language} & \textbf{Language Code\textsuperscript{\ddag}} & \textbf{Script} & \textbf{Language Family} & \textbf{Joshi Class\textsuperscript{$\dagger$}} & \textbf{Vitality} \\
        \midrule
            Afrikaans & afr & Latin & Germanic & 3 & MRL \\
            Akan & twi & Latin & Atlantic-Congo & 1 & LRL \\
            Amharic & amh & Amharic & Semitic & 2 & LRL \\
            Arabic & ara & Arabic & Semitic & 5 & HRL \\
            Azerbaijani & aze & Latin & Oghuz & 1 & LRL \\
            Bambara & bam & Latin & Mande & 1 & LRL \\
            Basque & eus & Latin & N/A & 4 & MRL \\
            Bengali & ben & Bengali & Indo-Iranian & 3 & MRL \\
            Bulgarian & bul & Cyrillic & Balto-Slavic & 3 & MRL \\
            Burmese & mya & Burmese & Sino-Tibetan & 1 & LRL \\
            Chinese & zho & Chinese & Sino-Tibetan & 5 & HRL \\
            Dholuo & luo & Latin & Nilo-Saharan & 0 & LRL \\
            Dutch & nld & Latin & Germanic & 4 & MRL \\
            Eastern Punjabi & pan & Gurmukhi & Indo-Iranian & 2 & LRL \\
            Estonian & est & Latin & Finnic & 3 & MRL \\
            Ewe & ewe & Latin & Atlantic-Congo & 1 & LRL \\
            Finnish & fin & Latin & Finnic & 4 & MRL \\
            Fon & fon & Latin & Atlantic-Congo & 0 & LRL \\
            French & fra & Latin & Romance & 5 & HRL \\
            Ganda & lug & Latin & Atlantic-Congo & 1 & LRL \\
            Georgian & kat & Georgian & Kartvelian & 3 & MRL \\
            German & deu & Latin & Germanic & 5 & HRL \\
            Greek & ell & Greek & Hellenic & 3 & MRL \\
            Gujarati & guj & Gujarati & Indo-Iranian & 1 & LRL \\
            Hausa & hau & Latin & Chadic & 2 & LRL \\
            Hebrew & heb & Hebrew & Semitic & 3 & MRL \\
            Hindi & hin & Devanagari & Indo-Iranian & 4 & MRL \\
            Hungarian & hun & Latin & Ugric & 4 & MRL \\
            Igbo & ibo & Latin & Atlantic-Congo & 1 & LRL \\
            Indonesian & ind & Latin & Malayic & 3 & MRL \\
            Italian & ita & Latin & Romance & 4 & MRL \\
            Japanese & jpn & Japanese & Japonic & 5 & HRL \\
            Javanese & jav & Latin & Javanic & 1 & LRL \\
            Kazakh & kaz & Cyrillic & Kipchak & 3 & MRL \\
            Kinyarwanda & kin & Latin & Atlantic-Congo & 1 & LRL \\
            Korean & kor & Korean & Koreanic & 4 & MRL \\
            Lingala & lin & Latin & Atlantic-Congo & 1 & LRL \\
            Lithuanian & lit & Latin & Balto-Slavic & 3 & MRL \\
            Malay & msa & Latin & Malayic & 3 & MRL \\
            Malayalam & mal & Malayalam & Southern Dravidian & 1 & LRL \\
            Marathi & mar & Devanagari & Indo-Iranian & 2 & LRL \\
            Mossi & mos & Latin & Gur & 0 & LRL \\
            Nyanja & nya & Latin & Benue-Congo & 1 & LRL \\
            Oromo & orm & Latin & Cushitic & 1 & LRL \\
            Persian & fas & Arabic & Indo-Iranian & 4 & MRL \\
            Polish & pol & Latin & Balto-Slavic & 4 & MRL \\
            Portuguese & por & Latin & Romance & 4 & MRL \\
            Romanian & ron & Latin & Romance & 3 & MRL \\
            Russian & rus & Cyrillic & Balto-Slavic & 4 & MRL \\
            Shona & sna & Latin & Atlantic-Congo & 1 & LRL \\
            Spanish & spa & Latin & Romance & 5 & HRL \\
            Swahili & swa & Latin & Atlantic-Congo & 2 & LRL \\
            Tagalog & tgl & Latin & Philippine & 3 & MRL \\
            Tamil & tam & Tamil & Southern Dravidian & 3 & MRL \\
            Telugu & tel & Telugu & South-Central Dravidian & 1 & LRL \\
            Thai & tha & Thai & Kra–Dai & 3 & MRL \\
            Tswana & tsn & Latin & Atlantic-Congo & 2 & LRL \\
            Turkish & tur & Latin & Oghuz & 4 & MRL \\
            Ukrainian & ukr & Cyrillic & Balto-Slavic & 3 & MRL \\
            Urdu & urd & Arabic & Indo-Iranian & 3 & MRL \\
            Vietnamese & vie & Latin & Austroasiatic & 4 & MRL \\
            Wolof & wol & Latin & Atlantic-Congo & 2 & LRL \\
            Xhosa & xho & Latin & Atlantic-Congo & 2 & LRL \\
            Yoruba & yor & Latin & Atlantic-Congo & 2 & LRL \\
            Zulu & zul & Latin & Atlantic-Congo & 2 & LRL \\
        \bottomrule
\caption{The 65 languages used for the realignment phase with their vitality class mapping. The language codes follow \textsuperscript{\ddag}ISO639-3 coding. Languages are mapped to their \textsuperscript{$\dagger$}rarity taxonomy based on ~\citet{joshi-etal-2020-state} vitality classes: Low Resource Language (LRL, 0-2), Medium Resource Language (MRL, 3-4), and High Resource Language (HRL, 5).}
\label{tab:languages}
\end{longtable}

\captionsetup[longtable]{skip=10pt}
\onecolumn
    \begin{longtable}{@{}p{0.3\linewidth}lp{0.5\linewidth}@{}}
\toprule
\textbf{Method} & \textbf{\#} & \textbf{Languages\textsuperscript{\ddag}} \\
\midrule
\textbf{Baseline} & & \\
\midrule
All 65 languages & 65 & 
    afr, amh, ara, aze, bul, ben, deu, ell, spa, est, eus, fas, fin, fra, \\ & & guj, heb, hin, hun, ind, ita, jpn, kat, kaz, kor, lit, mal, mar, msa, \\ & & mya, nld, pan, pol, por, ron, rus, tam, tha, tur, ukr, urd, vie, zho, \\ & & bam, ewe, fon, hau, ibo, kin, lin, lug, luo, mos, nya, gaz, sna, swh, \\ & & tsn, twi, wol, xho, yor, zul, jav, tgl, tel\\
    [6pt]
\addlinespace
Present in XTREME-R & 47 & 
    afr, ara, aze, bul, ben, deu, ell, spa, est, eus, fas, fin, fra, guj, \\ & & heb, hin, hun, ind, ita, jpn, jav, kat, kaz, kor, lit, mar, mal, msa, mya, \\ & & nld, pan, pol, por, ron, rus, swh, tam, tel, tgl, tha, tur, ukr, urd, \\ & & vie, wol, yor, zho\\
    [6pt]
\addlinespace
Present in Africa & 21 & 
    amh, bam, ewe, fon, hau, ibo, kin, lin, lug, luo, mos, nya, gaz, sna, \\ & & swh, tsn, twi, wol, xho, yor, zul \\

\midrule

\textbf{Featural Diversity} & & \\
\midrule
Most diverse from English & 5 & fon, kat, kaz, lin, gaz \\
& 10 & afr, ara, fon, kat, jpn, kaz, lin, gaz, sna, vie \\
& 20 & afr, ara, aze, eus, zho, fon, lug, kat, ell, hau, heb, ibo, jpn, kaz, \\ & & kor, lin, luo, gaz, sna, twi, vie, yor \\
& 40 & afr, ara, aze, eus, mya, zho, ewe, fon, fra, lug, kat, ell, hau, heb \\ & &
ibo, jpn, kaz, kin, kor, lin, msa, mal, mar, nya, gaz, fas, rus, sna \\ & &
spa, tgl, tam, tel, tha, tur, twi, urd, vie, xho, yor, zul
\\
[6pt]
\addlinespace
Least diverse from English & 5 & ita, por, ron, spa, ukr \\
& 10 & bul, deu, ell, spa, fra, ita, nld, por, ron, ukr \\
& 20 & bul, nld, est, fin, fra, deu, ell, guj, hin, hun, ita, lit, fas, pol \\ & & por, pan, ron, rus, spa, ukr\\
& 40 & amh, ara, aze, bam, eus, ben, bul, nld, est, fin, fra, deu, ell, guj \\ & & hau, heb, hin, hun, ind, ita, jav, lit, luo, mal, mar, mos, fas, pol \\ & & por, pan, ron, rus, spa, tgl, tam, tel, tur, ukr, urd, wol\\
\addlinespace
\midrule

\textbf{Phylogenetic Diversity} & & \\
\midrule
Most diverse families & 5 & kat, kaz, lin, gaz, vie \\
& 10 & ara, zho, kat, jpn, kaz, lin, msa, gaz, tam, vie \\
& 20 & ara, aze, eus, zho, fra, kat, ell, hau, jpn, kaz, kor, lin, luo, msa, \\ & & mar, gaz, rus, tam, tha, vie \\
& 25 & ara, aze, bam, eus, zho, fin, fra, kat, ell, hau, hun, jpn, kaz, kor \\ & & lin, luo, msa, mar, mos, gaz, rus, tam, tel, tha, vie \\
[6pt]
\addlinespace
Most diverse families within Indo-European & 5 & afr, nld, deu, ita, por \\
& 10 & afr, bul, nld, fra, deu, ita, por, ron, spa, ukr \\
& 20 & afr, ben, bul, nld, fra, deu, ell, guj, hin, ita, lit, mar, pol, por \\ & & pan, ron, rus, spa, ukr, urd \\

\midrule

\textbf{Script Diversity} & & \\
\midrule
Most diverse using distinct scripts & 5 & ara, kat, jpn, kaz, tha \\
& 10 & ara, mya, zho, kat, ell, heb, jpn, kaz, tam, tha \\
& 18 & amh, ara, ben, mya, zho, kat, ell, guj, heb, hin, jpn, kaz, kor, mal, \\ & & pan, tam, tel, tha \\[6pt]
\addlinespace
Most diverse using Latin scripts & 5 & aze, fon, lin, gaz, tgl \\
& 10 & afr, aze, eus, fon, lin, gaz, sna, tgl, twi, vie \\
& 20 & bam, nld, est, fin, fra, deu, hau, hun, ind, ita, jav, lit, luo, msa, \\ & & pol, por, ron, spa, tgl, wol \\ 
& 41 & afr, aze, bam, eus, nld, est, ewe, fin, fon, fra, lug, deu, hau, hun \\ & & ibo, ind, ita, jav, kin, lin, lit, luo, msa, mos, nya, gaz, pol, por \\ & & ron, sna, spa, swh, tgl, tsn, tur, twi, vie, wol, xho, yor, zul\\
[6pt]
\addlinespace
Least diverse using Latin scripts & 5 & nld, fra, deu, ita, por \\
& 10 & nld, est, fin, fra, deu, hun, ita, por, ron, spa \\
& 20 & afr, aze, eus, ewe, fon, fra, lug, lin, lit, msa, gaz, pol, sna, spa \\& & tgl, tur, twi, vie, yor, zul \\

\midrule
\textbf{Ablation Languages} & & \\
\midrule
Joshi Class = 2 (Random) & 10 & amh, aze, bam, ewe, fon, gaz, guj, hau, ibo, jav \\
Joshi Class = 2 (Most URIEL) & 10 & aze, mya, fon, kin, lin, mar, gaz, sna, tel, yor \\
Joshi Class = 2 (Most Family) & 10 & aze, mya, hau, jav, lin, luo, mar, mos, gaz, tel \\
Joshi Class = 2 (Most Script) & 10 & amh, mya, guj, lin, mal, mar, gaz, pan, tel, yor \\
[6pt]
Joshi Class = 3 (Random) & 17 & afr, bul, ben, ell, est, heb, ind, kat, kaz, lit, msa, \\ & & ron, tam, tgl, tha, ukr, urd \\
Joshi Class = 3 (Most URIEL) & 10 & afr, kat, ell, heb, kaz, msa, tgl, tam, tha, urd \\
Joshi Class = 3 (Most Family) & 10 & est, kat, ell, heb, kaz, lit, msa, tam, tha, urd \\
Joshi Class = 3 (Most Script) & 10 & ben, bul, kat, ell, heb, kaz, tam, tha, ukr, urd \\
[6pt]
Joshi Class = 3,5 (Random) & 23 & afr, ara, bul, ben, deu, ell, spa, est, eus, fas, fin, \\ & & fra, heb, hin, hun, ind, ita, jpn, kat, kaz, kor, lit, msa, \\ & & nld, pan, pol, por, ron, rus, tam, tgl, tha, tur, ukr, urd, vie, zho \\
Joshi Class = 3,5 (Most URIEL) & 10 & afr, ara, eus, zho, kat, jpn, kaz, msa, tam, vie \\
Joshi Class = 3,5 (Most Family) & 10 & ara, zho, kat, ell, jpn, kaz, msa, tam, tha, vie \\
Joshi Class = 3,5 (Most Script) & 10 & ara, zho, kat, ell, heb, jpn, kaz, kor, tam, tha \\
[6pt]
Joshi Class = 4,5 (Random) & 20 & ara, deu, spa, eus, fas, fin, fra, hin, hun, ita, jpn, \\ & & kor, nld, pol, por, rus, tam, tur, vie, zho \\
Joshi Class = 4,5 (Most URIEL) & 10 & ara, eus, zho, fra, jpn, kor, fas, rus, tur, vie \\
Joshi Class = 4,5 (Most Script) & 10 & ara, eus, fra, hin, jpn, kor, fas, rus, tur, vie \\
[6pt]
Seen by mBERT (Random) & 47 & afr, ara, aze, bul, ben, deu, ell, spa, est, eus, fas, \\ & & fin, fra, guj, heb, hin, hun, ind, ita, jpn, jav, kat, kaz, kor, \\ & & lit, mar, mal, msa, mya, nld, pan, pol, por, ron, rus, swh, \\ & & tam, tel, tgl, tha, tur, ukr, urd, vie, yor, zho \\
Seen by mBERT (Most URIEL) & 10 & afr, ara, zho, kat, jpn, kaz, swh, tam, vie, yor \\
Seen by mBERT (Most Family) & 10 & ara, aze, zho, kat, jpn, kaz, msa, tam, vie, yor \\
Seen by mBERT (Most Script) & 10 & ara, mya, zho, kat, ell, heb, jpn, kaz, tam, tha \\
[6pt]
Seen by XLM-R (Random) & 51 & afr, amh, ara, aze, bul, ben, deu, ell, spa, est, eus, \\ & & fas, fin, fra, gaz, guj, hau, heb, hin, hun, ind, ita, jpn, jav, \\ & & kat, kaz, kor, lit, mar, mal, msa, mya, nld, pan, pol, por, \\ & & ron, rus, swh, tam, tel, tgl, tha, tur, ukr, urd, vie, xho, zho \\
Seen by XLM-R (Most URIEL) & 10 & afr, ara, zho, kat, jpn, kaz, msa, gaz, swh, vie \\
Seen by XLM-R (Most Family) & 10 & ara, zho, kat, jpn, kaz, msa, gaz, tam, vie, xho \\
Seen by XLM-R (Most Script) & 10 & ara, mya, zho, kat, ell, heb, jpn, kaz, tam, tha \\
[6pt]
Unseen by mBERT (Random) & 34 & amh, bam, ewe, fon, gaz, hau, ibo, kin, lin, lug, \\ & & luo, mos, nya, sna, tsn, twi, wol, xho, yor, zul \\
Unseen by mBERT (Most URIEL) & 10 & amh, ewe, fon, hau, lin, luo, gaz, sna, twi, xho \\
Unseen by mBERT (Most Family) & 10 & amh, bam, fon, hau, lin, luo, mos, gaz, sna, twi \\
[6pt]
Unseen by XLM-R (Random) & 30 & bam, ewe, fon, ibo, kin, lin, lug, luo, mos, nya, \\ & & sna, tsn, twi, wol, yor, zul \\
Unseen by XLM-R (Most URIEL) & 10 & ewe, fon, lin, luo, mos, sna, twi, wol, yor, zul \\
Unseen by XLM-R (Most Family) & 10 & bam, fon, lin, luo, mos, sna, twi, wol, yor, zul \\

\bottomrule
\caption{List of languages used for each experiment and selection strategy. Language codes follow \textsuperscript{\ddag}ISO639-3 coding .}
\label{tab:experiment-languages}
\end{longtable}

\captionsetup[longtable]{skip=10pt}
\onecolumn
\begin{longtable}{@{}lccc|cccc@{}}
    \toprule
    & & & & \multicolumn{4}{c}{\textbf{Seen by}} \\
    \cmidrule(lr){5-8}
    \textbf{Language} & \textbf{NLI} & \textbf{PoS-tagging} & \textbf{NER} & \textbf{XLM-R} & \textbf{mBERT} & \textbf{OPUS-100} & \textbf{NLLB} \\ 
    \midrule
    \endfirsthead
    \multicolumn{8}{c}{\small\itshape (continued)}\\
    \toprule
    \textbf{Language} & \textbf{NLI} & \textbf{PoS-tagging} & \textbf{NER} & \textbf{XLM-R} & \textbf{mBERT} & \textbf{OPUS-100} & \textbf{NLLB}\\
    \midrule
    \endhead
        English (training) & 392,702 & 21,787 & 20,000 & \checkmark & \checkmark & \checkmark & \checkmark \\
        \midrule
        Afrikaans & - & 425 & 1000 & \checkmark & \checkmark & \checkmark & \checkmark \\
        Arabic & 5010 & 1680 & 10000 & \checkmark & \checkmark & \checkmark & \checkmark \\
        Azerbaijani & - & - & 1000 & \checkmark & \checkmark & \checkmark & \checkmark \\
        Basque & - & 1799 & 10000 & \checkmark & \checkmark & \checkmark & \checkmark \\
        Bengali & - & - & 1000 & \checkmark & \checkmark & \checkmark & \checkmark \\
        Bulgarian & 5010 & 1116 & 10000 & \checkmark & \checkmark & \checkmark & \checkmark \\
        Burmese & 5010 & - & 100 & \checkmark & \checkmark & \checkmark & \checkmark \\
        Chinese & 5010 & 3455 & 10000 & \checkmark & \checkmark & \checkmark & \checkmark \\
        Czech & - & 10159 & - & \checkmark & \checkmark & \checkmark & \checkmark \\
        Dutch & - & 1471 & 10000 & \checkmark & \checkmark & \checkmark & \checkmark \\
        English (evaluation) & 5010 & 5440 & 10000 & \checkmark & \checkmark & \checkmark & \checkmark \\
        Estonian & - & 4127 & 10000 & \checkmark & \checkmark & \checkmark & \checkmark \\
        Finnish & - & 6544 & 10000 & \checkmark & \checkmark & \checkmark & \checkmark \\
        French & 5010 & 7542 & 10000 & \checkmark & \checkmark & \checkmark & \checkmark \\
        Georgian & - & - & 10000 & \checkmark & \checkmark & \checkmark & \checkmark \\
        German & 5010 & 22358 & 10000 & \checkmark & \checkmark & \checkmark & \checkmark \\
        Greek & 5010 & 456 & 10000 & \checkmark & \checkmark & \checkmark & \checkmark \\
        Gujarati & - & - & 100 & \checkmark & \checkmark & \checkmark & \checkmark \\
        Hebrew & - & 491 & 10000 & \checkmark & \checkmark & \checkmark & \checkmark \\
        Hindi & 5010 & 2684 & 1000 & \checkmark & \checkmark & \checkmark & \checkmark \\
        Hungarian & - & 449 & 10000 & \checkmark & \checkmark & \checkmark & \checkmark \\
        Indonesian & 2984 & 1931 & 10000 & \checkmark & \checkmark & \checkmark & \checkmark \\
        Italian & - & 3518 & 10000 & \checkmark & \checkmark & \checkmark & \checkmark \\
        Japanese & - & 2365 & 10000 & \checkmark & \checkmark & \checkmark & \checkmark \\
        Javanese & - & - & 100 & \checkmark & \checkmark & \xmark & \checkmark \\
        Kazakh & - & 1047 & 1000 & \checkmark & \checkmark & \checkmark & \checkmark \\
        Korean & - & 4276 & 10000 & \checkmark & \checkmark & \checkmark & \checkmark \\
        Lithuanian & - & 739 & 10000 & \checkmark & \checkmark & \checkmark & \checkmark \\
        Malay & - & - & 1000 & \checkmark & \checkmark & \checkmark & \checkmark \\
        Malayalam & - & - & 1000 & \checkmark & \checkmark & \checkmark & \checkmark \\
        Marathi & - & 47 & 1000 & \checkmark & \checkmark & \checkmark & \checkmark \\
        Persian & - & 2055 & 10000 & \checkmark & \checkmark & \checkmark & \checkmark \\
        Polish & - & 4942 & 10000 & \checkmark & \checkmark & \checkmark & \checkmark \\
        Portuguese & - & 2680 & 10000 & \checkmark & \checkmark & \checkmark & \checkmark \\
        Eastern Punjabi & - & - & 100 & \checkmark & \checkmark & \checkmark & \checkmark \\
        Romanian & - & 2272 & 10000 & \checkmark & \checkmark & \checkmark & \checkmark \\
        Russian & 5010 & 8973 & 10000 & \checkmark & \checkmark & \checkmark & \checkmark \\
        Spanish & 5010 & 3147 & 10000 & \checkmark & \checkmark & \checkmark & \checkmark \\
        Tagalog & - & - & 1000 & \checkmark & \checkmark & \xmark & \checkmark \\
        Tamil & - & 654 & 1000 & \checkmark & \checkmark & \checkmark & \checkmark \\
        Telugu & - & 146 & 1000 & \checkmark & \checkmark & \checkmark & \checkmark \\
        Thai & 5010 & 1000 & 10000 & \checkmark & \checkmark & \checkmark & \checkmark \\
        Turkish & 5010 & 6647 & 10000 & \checkmark & \checkmark & \checkmark & \checkmark \\
        Ukrainian & - & 892 & 10000 & \checkmark & \checkmark & \checkmark & \checkmark \\
        Urdu & 5010 & 535 & 1000 & \checkmark & \checkmark & \checkmark & \checkmark \\
        Vietnamese & 5010 & 800 & 10000 & \checkmark & \checkmark & \checkmark & \checkmark \\
        \midrule
        Akan & - & 628 & 1211 & \xmark & \xmark & \xmark & \checkmark \\
        Amharic & 600 & - & - & \checkmark & \xmark & \checkmark & \checkmark \\
        Bambara & - & 619 & 1274 & \xmark & \xmark & \xmark & \checkmark \\
        Chichewa & - & - & 1785 & \xmark & \xmark & \xmark & \checkmark \\
        Dholuo & - & 606 & 1474 & \xmark & \xmark & \xmark & \checkmark \\
        Ewe & 600 & 582 & 1001 & \xmark & \xmark & \xmark & \checkmark \\
        Fon & - & 646 & 1228 & \xmark & \xmark & \xmark & \checkmark \\
        Ghomala & - & 599 & 966 & \xmark & \xmark & \xmark & \xmark \\
        Hausa & 600 & 601 & 1633 & \checkmark & \xmark & \checkmark & \checkmark \\
        Igbo & 600 & 642 & 2181 & \xmark & \xmark & \checkmark & \checkmark \\
        Kinyarwanda & 600 & 604 & 2235 & \xmark & \xmark & \checkmark & \checkmark \\
        Lingala & 600 & - & - & \xmark & \xmark & \xmark & \checkmark \\
        Ganda & 600 & 586 & 1412 & \xmark & \xmark & \xmark & \checkmark \\
        Mossi & - & 604 & 1294 & \xmark & \xmark & \xmark & \checkmark \\
        Naija & - & - & 1613 & \xmark & \xmark & \xmark & \xmark \\
        Oromo & 600 & - & - & \checkmark & \xmark & \checkmark & \xmark \\
        Setswana & - & 602 & 996 & \xmark & \xmark & \xmark & \checkmark \\
        Shona & 600 & 596 & 1773 & \xmark & \xmark & \xmark & \checkmark \\
        Southern Sotho & 600 & - & - & \xmark & \xmark & \xmark & \checkmark \\
        Swahili & 600 & 553 & 1883 & \checkmark & \checkmark & \xmark & \checkmark \\
        Wolof & 600 & 625 & 1312 & \xmark & \xmark & \xmark & \checkmark \\
        Xhosa & 600 & 601 & 1633 & \checkmark & \xmark & \checkmark & \checkmark \\
        Yoruba & 600 & 713 & 1964 & \xmark & \checkmark & \checkmark & \checkmark \\
        Zulu & 600 & 601 & 1670 & \xmark & \xmark & \checkmark & \checkmark \\
        \midrule
        Aymara & 750 & - & - & \xmark & \xmark & \xmark & \checkmark \\
        Asháninka & - & - & - & \xmark & \xmark & \xmark & \xmark \\
        Bribri & 750 & - & - & \xmark & \xmark & \xmark & \xmark \\
        Guaraní & 750 & - & - & \xmark & \xmark & \xmark & \checkmark \\
        Nahuatl & 750 & - & - & \xmark & \xmark & \xmark & \xmark \\
        Otomí & 750 & - & - & \xmark & \xmark & \xmark & \xmark \\
        Quechua & 750 & - & 100 & \xmark & \xmark & \xmark & \checkmark \\
        Rarámuri & 750 & - & - & \xmark & \xmark & \xmark & \xmark \\ 
        Shipibo-Konibo & 750 & - & - & \xmark & \xmark & \xmark & \xmark \\
        Wixárika & 750 & - & - & \xmark & \xmark & \xmark & \xmark \\
    \bottomrule
\caption{The size of the combined datasets. The table is split into 3 sections: 1) The original 44 languages used for realignment 2) African languages exclusive to AfriXNLI and MasakhaPOS, 3) South American languages exclusive to AmericasNLI.} \label{tab:dataset-statistics} \\
\end{longtable}
\captionsetup[longtable]{skip=10pt}
\onecolumn
\begin{longtable}{@{}p{0.3\linewidth}c|ccc@{}}
\toprule
\textbf{Method} & \textbf{\#Languages} & \textbf{NLI} & \textbf{PoS-Tagging} & \textbf{NER} \\
\midrule
\textbf{Baseline} & & & \\
\midrule
All 65 languages & 65 & \textbf{55.43 \scriptsize $\pm$ 0.29} & \textbf{66.87 \scriptsize $\pm$ 0.27} & \textbf{54.75 \scriptsize $\pm$ 1.02} \\
Present in XTREME-R & 47 & 53.51 \scriptsize $\pm$ 0.27 & 65.06 \scriptsize $\pm$ 0.38 & 52.54 \scriptsize $\pm$ 1.41 \\
Present in Africa & 21 & 54.94 \scriptsize $\pm$ 0.38 & 65.79 \scriptsize $\pm$ 0.18 & 53.72 \scriptsize $\pm$ 0.86 \\
Fine-tuning only & 0 & 53.12 \scriptsize $\pm$ 0.25 & 62.20 \scriptsize $\pm$ 0.78 & 52.25 \scriptsize $\pm$ 1.05 \\
\midrule

\textbf{Featural Diversity} & & & \\
\midrule
\textit{Most diverse from English} & 5 & 53.86 \scriptsize $\pm$ 0.24 & 64.08 \scriptsize $\pm$ 0.30 & 50.73 \scriptsize $\pm$ 0.73 \\
 & 10 & 54.40 \scriptsize $\pm$ 0.31 & 64.87 \scriptsize $\pm$ 0.27 & 53.64 \scriptsize $\pm$ 0.59 \\
 & 20 & 54.94 \scriptsize $\pm$ 0.31 & 65.66 \scriptsize $\pm$ 0.06 & 54.38 \scriptsize $\pm$ 1.05 \\
 & 40 & \textbf{55.94 \scriptsize $\pm$ 0.24} & \textbf{66.24 \scriptsize $\pm$ 0.07} & \textbf{54.82 \scriptsize $\pm$ 0.45} \\
[6pt]
\addlinespace
\textit{Least diverse from English} & 5 & 53.22 \scriptsize $\pm$ 0.17 & 62.57 \scriptsize $\pm$ 0.33 & 51.43 \scriptsize $\pm$ 1.41 \\
 & 10 & 53.26 \scriptsize $\pm$ 0.52 & 63.14 \scriptsize $\pm$ 0.60 & 52.70 \scriptsize $\pm$ 1.14 \\
 & 20 & 53.26 \scriptsize $\pm$ 0.12 & 63.33 \scriptsize $\pm$ 0.39 & 51.58 \scriptsize $\pm$ 0.69 \\
 & 40 & 53.69 \scriptsize $\pm$ 0.28 & 65.83 \scriptsize $\pm$ 0.08 & 52.85 \scriptsize $\pm$ 0.49 \\
\midrule

\textbf{Phylogenetic Diversity} & & & \\
\midrule
\textit{Most diverse families} & 5 & 54.03 \scriptsize $\pm$ 0.30 & 63.53 \scriptsize $\pm$ 0.06 & 49.69 \scriptsize $\pm$ 1.06 \\
 & 10 & 53.86 \scriptsize $\pm$ 0.25 & 63.85 \scriptsize $\pm$ 0.51 & 51.92 \scriptsize $\pm$ 0.82 \\
 & 20 & 53.98 \scriptsize $\pm$ 0.34 & 63.87 \scriptsize $\pm$ 0.18 & 53.98 \scriptsize $\pm$ 0.84 \\
 & 25 & \textbf{54.25 \scriptsize $\pm$ 0.19} & \textbf{65.13 \scriptsize $\pm$ 0.37} & \textbf{54.73 \scriptsize $\pm$ 0.85} \\
[6pt]
\addlinespace
\textit{Most diverse families within Indo-European} & 5 & 52.86 \scriptsize $\pm$ 0.35 & 61.89 \scriptsize $\pm$ 0.48 & 50.20 \scriptsize $\pm$ 0.68 \\
 & 10 & 53.18 \scriptsize $\pm$ 0.45 & 62.40 \scriptsize $\pm$ 0.49 & 52.52 \scriptsize $\pm$ 1.56 \\ 
 & 20 & 53.06 \scriptsize $\pm$ 0.29 & 63.59 \scriptsize $\pm$ 0.87 & 51.08 \scriptsize $\pm$ 0.80 \\
\midrule

\textbf{Script Diversity} & & & \\
\midrule
\textit{Most diverse scripts} & 5 & 52.80 \scriptsize $\pm$ 0.25 & 61.74 \scriptsize $\pm$ 0.66 & 51.19 \scriptsize $\pm$ 1.55 \\
 & 10 & 52.54 \scriptsize $\pm$ 0.34 & 62.11 \scriptsize $\pm$ 0.87 & 50.41 \scriptsize $\pm$ 1.14 \\
 & 18 & 52.69 \scriptsize $\pm$ 0.13 & 62.82 \scriptsize $\pm$ 0.74 & 50.59 \scriptsize $\pm$ 0.73 \\
[6pt]
\addlinespace
\textit{Most diverse using Latin script} & 5 & 53.94 \scriptsize $\pm$ 0.31 & 63.92 \scriptsize $\pm$ 0.38 & 51.38 \scriptsize $\pm$ 0.50 \\
 & 10 & 54.75 \scriptsize $\pm$ 0.13 & 64.95 \scriptsize $\pm$ 0.40 & 53.28 \scriptsize $\pm$ 0.73 \\
 & 20 & 53.89 \scriptsize $\pm$ 0.12 & 65.25 \scriptsize $\pm$ 0.15 & 53.10 \scriptsize $\pm$ 0.89 \\
 & 41 & \textbf{55.96 \scriptsize $\pm$ 0.15} & \textbf{67.23 \scriptsize $\pm$ 0.20} & \textbf{54.00 \scriptsize $\pm$ 0.48} \\
[6pt]
\addlinespace
\textit{Least diverse using Latin script} & 5 & 53.20 \scriptsize $\pm$ 0.56 & 61.36 \scriptsize $\pm$ 0.13 & 49.22 \scriptsize $\pm$ 0.47 \\
 & 10 & 53.14 \scriptsize $\pm$ 0.31 & 62.89 \scriptsize $\pm$ 0.51 & 51.04 \scriptsize $\pm$ 0.33 \\
 & 20 & 55.73 \scriptsize $\pm$ 0.19 & 66.05 \scriptsize $\pm$ 0.25 & 53.60 \scriptsize $\pm$ 0.91 \\
[6pt]
\addlinespace
\midrule

\textbf{Random Selection} & & & \\
\midrule
\textit{Random Seeded} & 5 & 54.21 \scriptsize $\pm$ 0.86 & 63.61 \scriptsize $\pm$ 0.55 & 51.76 \scriptsize $\pm$ 1.90 \\
 & 10 & 54.24 \scriptsize $\pm$ 0.88 & 64.78 \scriptsize $\pm$ 0.34 & 51.45 \scriptsize $\pm$ 0.51 \\
 & 20 & 54.64 \scriptsize $\pm$ 0.76 & 65.27 \scriptsize $\pm$ 0.42 & 52.92 \scriptsize $\pm$ 1.14 \\
 & 40 & \textbf{55.36 \scriptsize $\pm$ 0.24} & \textbf{66.06 \scriptsize $\pm$ 0.63} & \textbf{53.42 \scriptsize $\pm$ 0.48} \\

\bottomrule
\caption{Accuracy of XLM-R on NLI, PoS-Tagging, NER tasks. Results are averaged across 4 seeds along with the standard deviation.}
\label{tab:results-detailed-xlmr}
\end{longtable}

\captionsetup[longtable]{skip=10pt}
\onecolumn
\begin{longtable}{@{}lc|cccccc@{}}
\toprule
\textbf{Method} & \textbf{\#Languages} & \textbf{NLI} & \textbf{PoS-Tagging} & \textbf{NER} \\
\midrule

\multicolumn{5}{l}{\textbf{Most featural diversity from English}} \\
\midrule
Joshi Class = 2 & 10 &54.88 \scriptsize $\pm$ 0.18 & \textbf{64.72 \scriptsize $\pm$ 0.26} & 51.42 \scriptsize $\pm$ 0.30 \\
Joshi Class = 3 & 17 &53.06 \scriptsize $\pm$ 0.36 & 63.30 \scriptsize $\pm$ 0.64 & 50.17 \scriptsize $\pm$ 0.81 \\
Joshi Class = 3,4,5 & 37 &53.20 \scriptsize $\pm$ 0.47 & 63.35 \scriptsize $\pm$ 1.02 & 51.13 \scriptsize $\pm$ 1.74 \\
Joshi Class = 4,5 & 20 &53.24 \scriptsize $\pm$ 0.30 & 63.84 \scriptsize $\pm$ 0.52 & 50.66 \scriptsize $\pm$ 0.41 \\
Seen by mBERT & 47 &53.85 \scriptsize $\pm$ 0.32 & 64.50 \scriptsize $\pm$ 0.88 & \textbf{52.96 \scriptsize $\pm$ 1.14} \\
Seen by XLM-R & 51 &54.27 \scriptsize $\pm$ 0.28 & 64.39 \scriptsize $\pm$ 0.26 & 52.71 \scriptsize $\pm$ 0.87 \\
Unseen by mBERT & 34 &\textbf{55.14 \scriptsize $\pm$ 0.40} & 64.02 \scriptsize $\pm$ 0.23 & 52.63 \scriptsize $\pm$ 0.44 \\
Unseen by XLM-R & 30 &54.88 \scriptsize $\pm$ 0.34 & 64.70 \scriptsize $\pm$ 0.58 & 52.80 \scriptsize $\pm$ 0.73 \\
[6pt]
\addlinespace
\midrule

\multicolumn{5}{l}{\textbf{Most Phylogenetic Diversity}} \\
\midrule
Joshi Class = 2 & 10 &54.11 \scriptsize $\pm$ 0.60 & 63.96 \scriptsize $\pm$ 0.18 & 51.23 \scriptsize $\pm$ 0.68 \\
Joshi Class = 3 & 17 &52.96 \scriptsize $\pm$ 0.39 & 63.60 \scriptsize $\pm$ 0.49 & 50.73 \scriptsize $\pm$ 1.91 \\
Joshi Class = 3,4,5 & 37 &53.10 \scriptsize $\pm$ 0.29 & 63.68 \scriptsize $\pm$ 0.23 & 52.53 \scriptsize $\pm$ 1.67 \\
Seen by mBERT & 47 &53.57 \scriptsize $\pm$ 0.25 & 64.06 \scriptsize $\pm$ 0.67 & 52.52 \scriptsize $\pm$ 1.12 \\
Seen by XLM-R & 51 &54.42 \scriptsize $\pm$ 0.58 & 64.26 \scriptsize $\pm$ 0.17 & 53.34 \scriptsize $\pm$ 0.80 \\
Unseen by mBERT & 34 &\textbf{54.59 \scriptsize $\pm$ 0.37} & 64.02 \scriptsize $\pm$ 0.33 & \textbf{53.37 \scriptsize $\pm$ 0.82} \\
Unseen by XLM-R & 30 &\textbf{54.59 \scriptsize $\pm$ 0.38} & \textbf{65.04 \scriptsize $\pm$ 0.47} & 52.54 \scriptsize $\pm$ 0.97 \\
[6pt]
\addlinespace
\midrule

\multicolumn{5}{l}{\textbf{Most Script Diversity}} \\
\midrule
Joshi Class = 2 & 10 & \textbf{54.20 \scriptsize $\pm$ 0.17} & \textbf{63.92 \scriptsize $\pm$ 0.28} & 50.50 \scriptsize $\pm$ 0.66 \\
Joshi Class = 3 & 17 & 52.67 \scriptsize $\pm$ 0.18 & 61.73 \scriptsize $\pm$ 1.00 & 49.46 \scriptsize $\pm$ 0.31 \\
Joshi Class = 3,4,5 & 37 & 52.60 \scriptsize $\pm$ 0.19 & 62.46 \scriptsize $\pm$ 0.57 & \textbf{52.24 \scriptsize $\pm$ 1.45} \\
Joshi Class = 4,5 & 20 & 53.16 \scriptsize $\pm$ 0.13 & 63.54 \scriptsize $\pm$ 0.35 & 51.61 \scriptsize $\pm$ 0.35 \\
Seen by mBERT & 47 & 52.73 \scriptsize $\pm$ 0.16 & 61.95 \scriptsize $\pm$ 0.37 & 49.73 \scriptsize $\pm$ 0.79 \\
Seen by XLM-R & 51 & 52.73 \scriptsize $\pm$ 0.16 & 61.95 \scriptsize $\pm$ 0.37 & 49.73 \scriptsize $\pm$ 0.79 \\
[6pt]
\addlinespace
\midrule

\multicolumn{5}{l}{\textbf{Random Seeded}} \\
\midrule
Joshi Class = 2 & 10 &\textbf{55.03 \scriptsize $\pm$ 0.65} & 65.06 \scriptsize $\pm$ 0.53 & 52.46 \scriptsize $\pm$ 0.97 \\
Joshi Class = 3 & 17 &53.28 \scriptsize $\pm$ 0.25 & 62.94 \scriptsize $\pm$ 0.49 & 50.31 \scriptsize $\pm$ 2.05 \\
Joshi Class = 3,4,5 & 37 &53.36 \scriptsize $\pm$ 0.21 & 63.54 \scriptsize $\pm$ 0.40 & 51.51 \scriptsize $\pm$ 0.70 \\
Joshi Class = 4,5 & 20 &53.30 \scriptsize $\pm$ 0.15 & 63.69 \scriptsize $\pm$ 0.14 & 50.67 \scriptsize $\pm$ 1.50 \\
Seen by mBERT & 47 &53.43 \scriptsize $\pm$ 0.37 & 63.67 \scriptsize $\pm$ 0.81 & 51.25 \scriptsize $\pm$ 0.40 \\
Seen by XLM-R & 51 &53.86 \scriptsize $\pm$ 0.48 & 63.84 \scriptsize $\pm$ 0.62 & 51.08 \scriptsize $\pm$ 0.65 \\
Unseen by mBERT & 34 &54.45 \scriptsize $\pm$ 0.87 & 64.55 \scriptsize $\pm$ 0.25 & 52.48 \scriptsize $\pm$ 1.19 \\
Unseen by XLM-R & 30 &54.87 \scriptsize $\pm$ 0.20 & \textbf{65.11 \scriptsize $\pm$ 0.29} & \textbf{52.62 \scriptsize $\pm$ 0.53} \\

\bottomrule
\caption{Ablation studies: Accuracy of mBERT on NLI, POS-Tagging, NER tasks. Results are averaged across 4 seeds along with the standard deviation.}
\label{tab:results-ablation-xlmr}
\end{longtable}

\captionsetup[longtable]{skip=10pt}
\onecolumn
\begin{longtable}{@{}p{0.3\linewidth}c|ccc@{}}
\toprule
\textbf{Method} & \textbf{\#Languages} & \textbf{NLI} & \textbf{PoS-Tagging} & \textbf{NER} \\
\midrule
\textbf{Baseline} & & & \\
\midrule
All 65 languages & 65 & 59.43 \scriptsize $\pm$ 0.17 & 69.14 \scriptsize $\pm$ 0.24 & 57.07 \scriptsize $\pm$ 0.86 \\
Present in XTREME-R & 47 & \textbf{59.44 \scriptsize $\pm$ 0.39} & 67.32 \scriptsize $\pm$ 0.59 & \textbf{57.24 \scriptsize $\pm$ 0.73} \\
Present in Africa & 21 & 59.90 \scriptsize $\pm$ 0.33 & \textbf{69.92 \scriptsize $\pm$ 0.15} & 54.10 \scriptsize $\pm$ 1.14 \\
Fine-tuning only & 0 & 58.61 \scriptsize $\pm$ 0.10 & 65.98 \scriptsize $\pm$ 0.73 & 51.09 \scriptsize $\pm$ 0.96 \\
\midrule

\textbf{Featural Diversity} & & & \\
\midrule
\textit{Most diverse from English} & 5 & 58.97 \scriptsize $\pm$ 0.28 & 67.99 \scriptsize $\pm$ 0.37 & 53.39 \scriptsize $\pm$ 1.17 \\
 & 10 & 58.87 \scriptsize $\pm$ 0.19 & 68.48 \scriptsize $\pm$ 0.43 & 56.11 \scriptsize $\pm$ 0.77 \\
 & 20 & 59.27 \scriptsize $\pm$ 0.07 & 68.57 \scriptsize $\pm$ 0.40 & 56.51 \scriptsize $\pm$ 1.16 \\
 & 40 & \textbf{59.64 \scriptsize $\pm$ 0.24} & \textbf{68.89 \scriptsize $\pm$ 0.26} & \textbf{56.39 \scriptsize $\pm$ 0.98} \\
[6pt]
\addlinespace
\textit{Least diverse from English} & 5 & 58.59 \scriptsize $\pm$ 0.17 & 66.04 \scriptsize $\pm$ 0.79 & 52.80 \scriptsize $\pm$ 0.58 \\
 & 10 & 58.53 \scriptsize $\pm$ 0.09 & 65.99 \scriptsize $\pm$ 0.80 & 52.87 \scriptsize $\pm$ 1.47 \\
 & 20 & 58.74 \scriptsize $\pm$ 0.11 & 67.10 \scriptsize $\pm$ 0.25 & 53.96 \scriptsize $\pm$ 1.34 \\
 & 40 & 58.75 \scriptsize $\pm$ 0.17 & 68.10 \scriptsize $\pm$ 0.36 & 56.21 \scriptsize $\pm$ 0.42 \\
\midrule

\textbf{Phylogenetic Diversity} & & & \\
\midrule
\textit{Most diverse families} & 5 & 58.81 \scriptsize $\pm$ 0.20 & 67.20 \scriptsize $\pm$ 0.37 & 51.63 \scriptsize $\pm$ 0.72 \\
 & 10 & 58.91 \scriptsize $\pm$ 0.29 & 67.31 \scriptsize $\pm$ 0.22 & 53.81 \scriptsize $\pm$ 1.04 \\
 & 20 & \textbf{59.15 \scriptsize $\pm$ 0.31} & 66.64 \scriptsize $\pm$ 0.10 & \textbf{55.53 \scriptsize $\pm$ 0.77} \\
 & 25 & 59.06 \scriptsize $\pm$ 0.12 & \textbf{67.87 \scriptsize $\pm$ 0.26} & 54.99 \scriptsize $\pm$ 0.66 \\
[6pt]
\addlinespace
\textit{Most diverse families within Indo-European} & 5 & 58.74 \scriptsize $\pm$ 0.30 & 66.39 \scriptsize $\pm$ 0.94 & 51.90 \scriptsize $\pm$ 1.23 \\
 & 10 & 58.59 \scriptsize $\pm$ 0.24 & 67.14 \scriptsize $\pm$ 0.64 & 53.81 \scriptsize $\pm$ 2.65 \\
 & 20 & 58.92 \scriptsize $\pm$ 0.10 & 67.15 \scriptsize $\pm$ 0.33 & 52.35 \scriptsize $\pm$ 1.42 \\
\midrule

\textbf{Script Diversity} & & & \\
\midrule
\textit{Most diverse scripts} & 5 & 58.76 \scriptsize $\pm$ 0.03 & 67.26 \scriptsize $\pm$ 0.46 & 49.89 \scriptsize $\pm$ 1.18 \\
 & 10 & 58.70 \scriptsize $\pm$ 0.27 & 67.04 \scriptsize $\pm$ 0.73 & 51.40 \scriptsize $\pm$ 1.97 \\
 & 18 & 58.77 \scriptsize $\pm$ 0.20 & 67.04 \scriptsize $\pm$ 0.92 & 50.83 \scriptsize $\pm$ 1.50 \\
[6pt]
\addlinespace
\textit{Most diverse using Latin script} & 5 & 58.75 \scriptsize $\pm$ 0.35 & 67.88 \scriptsize $\pm$ 0.26 & 53.74 \scriptsize $\pm$ 0.80 \\
 & 10 & 59.15 \scriptsize $\pm$ 0.30 & 68.11 \scriptsize $\pm$ 0.15 & 56.27 \scriptsize $\pm$ 0.34 \\
 & 20 & 58.75 \scriptsize $\pm$ 0.23 & 67.75 \scriptsize $\pm$ 0.09 & 55.47 \scriptsize $\pm$ 1.45 \\
 & 41 & \textbf{59.88 \scriptsize $\pm$ 0.06} & \textbf{69.62 \scriptsize $\pm$ 0.30} & \textbf{56.94 \scriptsize $\pm$ 0.44} \\
[6pt]
\addlinespace
\textit{Least diverse using Latin script} & 5 & 58.86 \scriptsize $\pm$ 0.28 & 65.49 \scriptsize $\pm$ 1.16 & 51.00 \scriptsize $\pm$ 1.02 \\
 & 10 & 58.77 \scriptsize $\pm$ 0.22 & 65.82 \scriptsize $\pm$ 1.30 & 53.27 \scriptsize $\pm$ 1.10 \\
 & 20 & 59.75 \scriptsize $\pm$ 0.04 & 68.65 \scriptsize $\pm$ 0.29 & 56.44 \scriptsize $\pm$ 0.22 \\
[6pt]
\addlinespace
\midrule

\textbf{Random Selection} & & & \\
\midrule
\textit{Random Seeded} & 5 & \textbf{59.49 \scriptsize $\pm$ 0.44} & 67.46 \scriptsize $\pm$ 1.51 & 54.18 \scriptsize $\pm$ 1.44 \\
 & 10 & 59.13 \scriptsize $\pm$ 0.50 & 67.91 \scriptsize $\pm$ 0.78 & 54.29 \scriptsize $\pm$ 1.59 \\
 & 20 & 59.27 \scriptsize $\pm$ 0.54 & 68.26 \scriptsize $\pm$ 0.46 & \textbf{56.66 \scriptsize $\pm$ 1.18} \\
 & 40 & \textbf{59.49 \scriptsize $\pm$ 0.20} & \textbf{68.69 \scriptsize $\pm$ 0.32} & 56.04 \scriptsize $\pm$ 0.73 \\

\bottomrule
\caption{Accuracy of mBERT on NLI, PoS-Tagging, NER tasks. Results are averaged across 4 seeds along with the standard deviation.}
\label{tab:results-detailed-mbert}
\end{longtable}

\captionsetup[longtable]{skip=10pt}
\onecolumn
\begin{longtable}{@{}lc|cccccc@{}}
\toprule
\textbf{Method} & \textbf{\#Languages} & \textbf{NLI} & \textbf{PoS-Tagging} & \textbf{NER} \\
\midrule

\multicolumn{5}{l}{\textbf{Most featural diversity from English}}\\
\midrule
Joshi Class = 2 & 10 &59.49 \scriptsize $\pm$ 0.25 & \textbf{68.83 \scriptsize $\pm$ 0.40} & 52.75 \scriptsize $\pm$ 1.66 \\
Joshi Class = 3 & 17 &59.00 \scriptsize $\pm$ 0.21 & 67.46 \scriptsize $\pm$ 0.36 & 51.90 \scriptsize $\pm$ 2.04 \\
Joshi Class = 3,4,5 & 37 &58.79 \scriptsize $\pm$ 0.15 & 67.13 \scriptsize $\pm$ 0.12 & 53.95 \scriptsize $\pm$ 1.32 \\
Joshi Class = 4,5 & 20 &58.75 \scriptsize $\pm$ 0.24 & 67.27 \scriptsize $\pm$ 0.37 & 54.26 \scriptsize $\pm$ 1.83 \\
Seen by mBERT & 47 &59.04 \scriptsize $\pm$ 0.38 & 68.04 \scriptsize $\pm$ 0.55 & \textbf{55.39 \scriptsize $\pm$ 0.98} \\
Seen by XLM-R & 51 &59.00 \scriptsize $\pm$ 0.38 & 67.58 \scriptsize $\pm$ 0.54 & 52.55 \scriptsize $\pm$ 1.75 \\
Unseen by mBERT & 34 &\textbf{59.86 \scriptsize $\pm$ 0.22} & 68.25 \scriptsize $\pm$ 0.30 & 54.70 \scriptsize $\pm$ 0.57 \\
Unseen by XLM-R & 30 &59.68 \scriptsize $\pm$ 0.44 & 68.62 \scriptsize $\pm$ 0.22 & 54.78 \scriptsize $\pm$ 1.39 \\
[6pt]
\addlinespace
\midrule

\multicolumn{5}{l}{\textbf{Most Phylogenetic Diversity}} \\
\midrule
Joshi Class = 2 & 10 &59.01 \scriptsize $\pm$ 0.12 & 67.59 \scriptsize $\pm$ 0.19 & 53.02 \scriptsize $\pm$ 1.29 \\
Joshi Class = 3 & 17 &58.88 \scriptsize $\pm$ 0.21 & 67.62 \scriptsize $\pm$ 0.34 & 51.15 \scriptsize $\pm$ 0.83 \\
Joshi Class = 3,4,5 & 37 &58.63 \scriptsize $\pm$ 0.20 & 67.91 \scriptsize $\pm$ 0.43 & 52.44 \scriptsize $\pm$ 0.68 \\
Seen by mBERT & 47 &58.79 \scriptsize $\pm$ 0.20 & 68.38 \scriptsize $\pm$ 0.25 & \textbf{56.23 \scriptsize $\pm$ 0.43} \\
Seen by XLM-R & 51 &59.07 \scriptsize $\pm$ 0.17 & 67.99 \scriptsize $\pm$ 0.22 & 55.08 \scriptsize $\pm$ 1.52 \\
Unseen by mBERT & 34 &\textbf{59.77 \scriptsize $\pm$ 0.10} & 68.58 \scriptsize $\pm$ 0.27 & 54.10 \scriptsize $\pm$ 1.09 \\
Unseen by XLM-R & 30 &59.34 \scriptsize $\pm$ 0.37 & \textbf{68.73 \scriptsize $\pm$ 0.13} & 53.34 \scriptsize $\pm$ 0.98 \\
[6pt]
\addlinespace
\midrule

\multicolumn{5}{l}{\textbf{Most Script Diversity}} \\
\midrule
Joshi Class = 2 & 10 & \textbf{59.11 \scriptsize $\pm$ 0.23} & 67.48 \scriptsize $\pm$ 0.15 & \textbf{54.33 \scriptsize $\pm$ 0.30} \\
Joshi Class = 3 & 17 & 58.47 \scriptsize $\pm$ 0.21 & 67.48 \scriptsize $\pm$ 0.35 & 50.89 \scriptsize $\pm$ 1.63 \\
Joshi Class = 3,4,5 & 37 & 58.66 \scriptsize $\pm$ 0.22 & \textbf{67.52 \scriptsize $\pm$ 0.28} & 51.64 \scriptsize $\pm$ 0.85 \\
Joshi Class = 4,5 & 20 & 58.65 \scriptsize $\pm$ 0.16 & 67.07 \scriptsize $\pm$ 0.29 & 53.26 \scriptsize $\pm$ 0.61 \\
Seen by mBERT & 47 & 58.67 \scriptsize $\pm$ 0.20 & 66.91 \scriptsize $\pm$ 1.18 & 50.46 \scriptsize $\pm$ 0.65 \\
Seen by XLM-R & 51 & 58.67 \scriptsize $\pm$ 0.20 & 66.91 \scriptsize $\pm$ 1.18 & 50.46 \scriptsize $\pm$ 0.65 \\
[6pt]
\addlinespace
\midrule

\multicolumn{5}{l}{\textbf{Random Seeded}} \\
\midrule
Joshi Class = 2 & 10 &\textbf{59.93 \scriptsize $\pm$ 0.05} & 68.84 \scriptsize $\pm$ 0.68 & \textbf{54.54 \scriptsize $\pm$ 1.80} \\
Joshi Class = 3 & 17 &58.95 \scriptsize $\pm$ 0.19 & 67.05 \scriptsize $\pm$ 0.79 & 51.00 \scriptsize $\pm$ 0.98 \\
Joshi Class = 3,4,5 & 37 &58.80 \scriptsize $\pm$ 0.24 & 67.16 \scriptsize $\pm$ 0.43 & 53.80 \scriptsize $\pm$ 1.23 \\
Joshi Class = 4,5 & 20 &58.69 \scriptsize $\pm$ 0.29 & 67.47 \scriptsize $\pm$ 0.70 & 54.38 \scriptsize $\pm$ 1.82 \\
Seen by mBERT & 47 &58.74 \scriptsize $\pm$ 0.13 & 67.39 \scriptsize $\pm$ 0.90 & 52.39 \scriptsize $\pm$ 2.13 \\
Seen by XLM-R & 51 &58.88 \scriptsize $\pm$ 0.21 & 67.10 \scriptsize $\pm$ 0.44 & 53.26 \scriptsize $\pm$ 2.65 \\
Unseen by mBERT & 34 &59.78 \scriptsize $\pm$ 0.38 & 69.01 \scriptsize $\pm$ 0.23 & 53.26 \scriptsize $\pm$ 1.33 \\
Unseen by XLM-R & 30 &59.82 \scriptsize $\pm$ 0.26 & \textbf{69.09 \scriptsize $\pm$ 0.14} & 53.54 \scriptsize $\pm$ 1.42 \\

\bottomrule
\caption{Ablation studies: Accuracy of XLM-R on NLI, PoS-Tagging, NER tasks. Results are averaged across 4 seeds along with the standard deviation.}
\label{tab:results-ablation-mbert}
\end{longtable}

\end{document}